\documentclass[letterpaper, 10 pt, conference]{ieeeconf}  % Comment this line out if you need a4paper

\IEEEoverridecommandlockouts                              % This command is only needed if 
\usepackage{amssymb}  % assumes amsmath package installed
\usepackage{graphics} % for pdf, bitmapped graphics files
\usepackage{graphicx}
\usepackage{hyperref}
\usepackage[tight,footnotesize]{subfigure}
\long\def\invis#1{}
\usepackage{cite}
\usepackage{url}
\usepackage{hyphenat}
\usepackage{amsmath}
\usepackage{xcolor}
\usepackage{siunitx}
\usepackage{xspace}
\usepackage{mathtools}
\usepackage{gensymb}
\usepackage{afterpage}

\newcommand\fig[1]{Figure~\ref{#1}}
\newcommand\tab[1]{Table~\ref{#1}}

\newcommand\etal{\textit{et al.\xspace}}

\usepackage{soul}

\setstcolor{blue}

\def\floatsep{4.0pt}
\def\textfloatsep{0.0pt}
\def\dblfloatsep{4.0pt}
\def\dbltextfloatsep{4.0pt}

\usepackage{fp}
\FPset{\pb}{0}
\newcommand{\pagebudget}[1]{}
\newcommand{\showtotalpagebudget}{}

\title{\LARGE \bf AquaVis: A Perception-Aware Autonomous\\Navigation Framework for Underwater Vehicles}

\author{Marios Xanthidis,$^1$ Michail Kalaitzakis,$^2$ Nare Karapetyan,$^1$ James Johnson,$^1$\\Nikolaos Vitzilaios,$^2$ Jason M. O'Kane,$^1$ and Ioannis Rekleitis$^1$%\vspace{-0.2in}% <-this % stops a space
\thanks{$^1$ M. Xanthidis, N. Karapetyan, A. Johnson, J. M. O'Kane, and I. Rekleitis are with the Department of Computer Science and Engineering, University of South Carolina, Columbia, SC, USA. {\tt\small [mariosx,nare,jvj1]@email.sc.edu, [jokane,yiannisr]@cse.sc.edu}}%
\thanks{$^2$M. Kalaitzakis and N. Vitzilaios are with the Department of Mechanical Engineering, University of South Carolina, Columbia, SC, USA. {\tt\small michailk@email.sc.edu,vitzilaios@sc.edu}}%
\thanks{This work was made possible through the generous support of  National Science Foundation grants (NSF   1659514, 1849291, 1943205,  2024741).}
}
\begin{document}

\maketitle
\thispagestyle{empty}
\pagestyle{empty}

%%%%%%%%%%%%%%%%%%%%%%%%%%%%%%%%%%%%%%%%%%%%%%%%%%%%%%%%%%%%%%%%%%%%%%%%%%%%%%%%
\begin{abstract}
Visual monitoring operations underwater require both observing the objects of interest in close-proximity, and tracking the few feature-rich areas necessary for state estimation.
This paper introduces the first navigation framework, called AquaVis, that produces on-line visibility-aware motion plans that enable Autonomous Underwater Vehicles (AUVs) to track multiple visual objectives with an arbitrary camera configuration in real-time.
Using the proposed pipeline, AUVs can efficiently move in 3D, reach their goals while avoiding obstacles safely, and maximizing the visibility of multiple objectives along the path within a specified proximity. The method is sufficiently fast to be executed in real-time and is suitable for single or multiple camera configurations.\invis{Testing utilizing the Aqua2 underwater robot~\cite{dudek2007aqua} in simulation} Experimental results show the significant improvement on tracking multiple automatically-extracted points of interest, with low computational overhead and fast re-planning times. 

Accompanying short video: \href{https://www.youtube.com/watch?v=JKO_bbrIZyU}{\textcolor{blue}{https://youtu.be/JKO\_bbrIZyU}}
\end{abstract}

%%%%%%%%%%%%%%%%%%%%%%%%%%%%%%%%%%%%%%%%%%%%%%%%%%%%%%%%%%%%%%%%%%%%%%%%%%%%%%%%

\section{Introduction}\pagebudget{1}
\label{intro}
%\vspace{-0.1in}
%3D navigation for a very agile robot using vision-based state estimation, and complex dynamics,

Autonomous underwater monitoring and navigation can be very hard for a variety of reasons.
For example, the robot must move safely, avoiding obstacles and staying at depth. Planning and executing such motions can be particularly challenging for an AUV moving in three dimensions, with complex dynamics that have not been adequately modeled. Our previous work introduced AquaNav~\cite{xanthidis2020navigation}, which robustly solved these problems for very challenging environments, in simulation, in-pool, and open water conditions.
Additionally, as a planning problem, visual monitoring of unknown underwater 3D environments in real-time is very challenging, due to the dimensionality of the problem, and the constraints introduced by the limited cameras' range and field of view. 

Furthermore, though visual data is generally utilized for state estimation, underwater environments tend to produce very noisy images due to lack of color saturation, insufficient illumination, and color attenuation.  Moreover, good visual features are often concentrated on few nearby objects; while much of the visible terrain 
% in marine environments 
has few features.
As a result, state-of-the-art methods fail to provide robust state estimation for the robot~\cite{joshi2019experimental,QuattriniLiISERVO2016}, although previous work has addressed this problem by providing a very capable SLAM framework called SVIn~\cite{RahmanICRA2018,RahmanIROS2019a}, under the assumption that an adequate number of high-quality features are visible throughout the path.
However, even the most capable vision-based SLAM systems have trouble tracking the state when facing featureless homogeneous surfaces, turbidity, or the open (blue) water conditions that dominate the underwater domain. 

% \st{, but can experience state estimation errors when portions of its plans lead the robot \invis{without visibility of useful} through areas deprived of features for an extended period.}

\begin{figure}[t]
\leavevmode
\begin{center}
   \begin{tabular}{c}
    %  \subfigure[]{\includegraphics[width=0.7\textwidth, clip=true, trim={0.0in 0.0in 0.0in 0.0in}]{figures/nav_vs_vis1}}\\
     \subfigure[]{\includegraphics[width=0.75\columnwidth, clip=true, trim={0.0in 0.0in 0.0in 0.0in}]{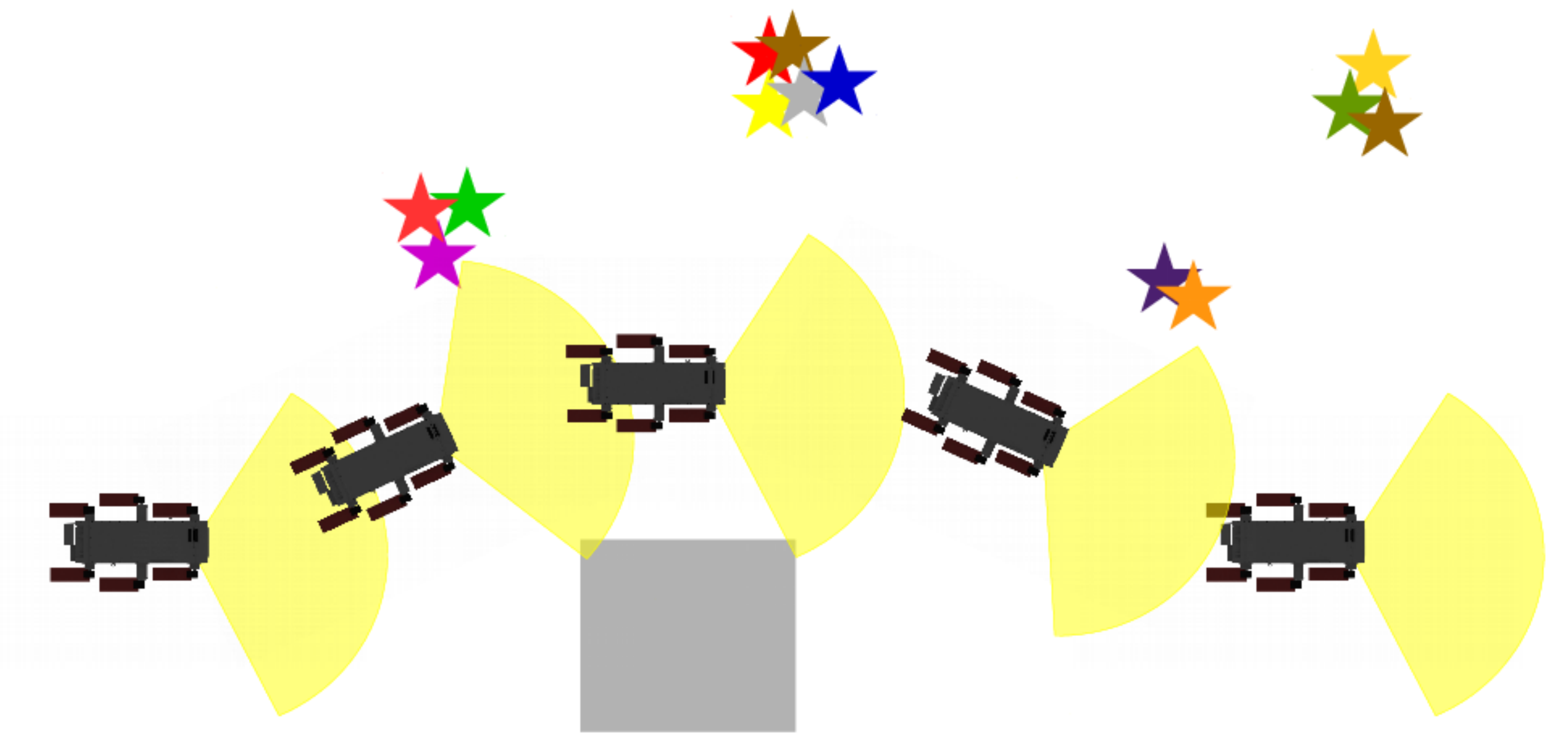}}\\
     \subfigure[]{\includegraphics[width=0.75\columnwidth, clip=true, trim={0.0in 0.0in 0.0in 0.0in}]{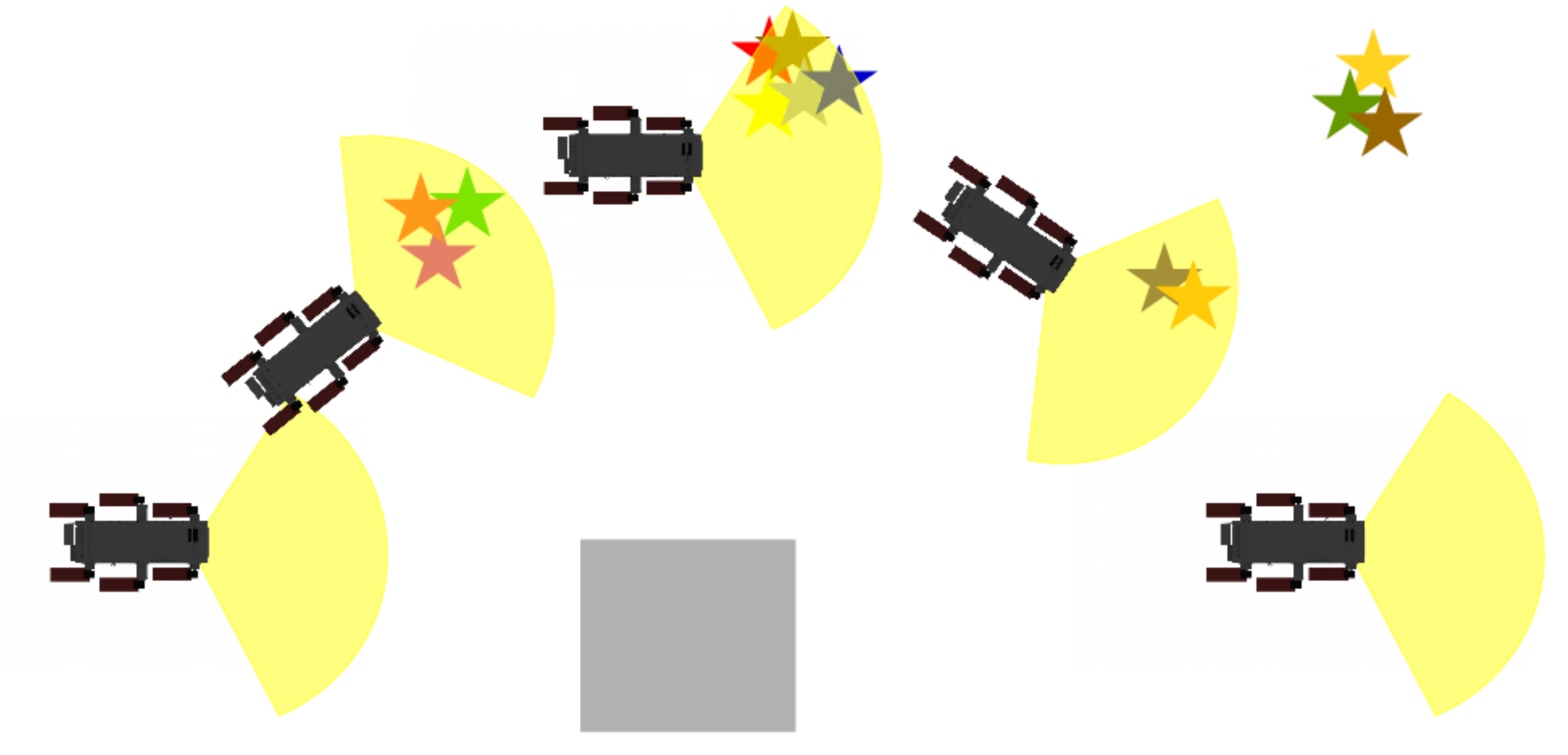}}
    \end{tabular}
%vspace{-0.1in}
\caption{An environment with obstacles (grey) and feature-rich visual objectives indicated with stars. (a) AquaNav,  considers only avoiding obstacles and minimizing the path length. (b) AquaVis, the method introduced here, navigates the robot safely by avoiding obstacles, while at the same time observing nearby visual objectives.}
\label{fig:nav_vis_dif}
\end{center}
\end{figure}

% Such failure cases, as indicated in~\cite{xanthidis2020navigation}, were common since often the optimal way for the robot to avoid detected obstacles, involve motions that steer the robot, and its cameras, away from these feature rich obstacles. 
\invis{Interestingly, Aqua2 following the decoupled approach of AquaNav, often avoids not only collisions with perceived objects, but also perceiving these objects that offered the best --- or even the only --- features essential for localization. }
For robust underwater navigation it is highly important to combine perception and motion planning, in order to avoid the obstacles, but also keep feature-rich objects in the cameras' field of view.
Bringing perception and motion planning closer not only assists state-estimation, but also produces trajectories that track and monitor points of interest, such as fish, corals, and structures. 
Such behavior is preferred for exploration and monitoring strategies that should collect diverse and meaningful-to-humans information~\cite{girdhar2015unsupervised,bourque2000automated}.
% Observing such objects of interest is not our primary focus --- which is the view of the typical coverage problem~\cite{rekleitis2008efficient,choset2001coverage,karapetyan2017efficient, karapetyan2019riverine} --- it is a secondary objective with the primary remaining the efficient and safe navigation to the specified goal position. 
To this end, we propose a novel framework called AquaVis, whose objective is to generate motions that enable the robot not only to move efficiently and avoid obstacles safely, but also to observe areas of interest, that could be extracted automatically. The difference is illustrated in Figure~\ref{fig:nav_vis_dif}.

This is achieved ---utilizing the\invis{high level of extensibility} flexibility of the AquaNav framework~\cite{xanthidis2020navigation}---
by introducing two novel cost functions in the optimization process during planning, to direct the robot to observe specific points of interest while avoiding the obstacles and respecting the kinematics of the robot. An analysis of the produced trajectories demonstrate object tracking with the desired proximity and safe navigation around obstacles.

% In the ideal case a solution would be a sequence of motions where if followed by the robot, the goal is reached, collisions are being avoided, the path length is kept minimal and during any transition, at least one point of interest is being observed at all times.  

The specific contributions of this paper are the following:
\begin{enumerate}
    \item A novel and robust framework, called AquaVis, for autonomous 3D navigation for the Aqua2 robot~\cite{dudek2007aqua}, surpassing AquaNav's capabilities by improving the perception capabilities.
    \item A novel formulation of perception-aware navigation for mobile robots with an arbitrary number of cameras, tracking multiple visual objectives, moving in 3D. We also show how visual objectives could be extracted automatically, from perceived point-clouds, to assist state estimation.
\invis{    \item Experimental results evaluating AquaVis on underwater navigation tasks with an AUV equipped with forward looking cameras, demonstrating that AquaVis enables the robot to:
        (i) track objects from a desired proximity, dealing effectively with turbidity; and
        (ii) safely navigate through challenging environments; inheriting properties of AquaNav.
        %(iii) being high-level, to allows modifications, adjustments, and exhibits sophisticated performance.  \textcolor{(This seems like a difficult claim to evaluate.  Do we need it?)}
        }
\end{enumerate}

\section{Related Work}\pagebudget{1}
\label{related_work}
% \vspace{-0.1in}
The problem of actively planning and executing motions to improve state estimation performance, also known as Active SLAM, was first introduced by Feder \etal~\cite{feder1999adaptive} in late 90's, in the context of exploration. 
% A framework is introduced that made local decisions to improve pose estimates during mapping, based on uniquely identified landmarks. 
Stachniss and Burgard~\cite{stachniss2004exploration} provided a method that improved localization using SLAM, by attempting loop-closing.

Makarenko \etal~\cite{makarenko2002experiment} employed a laser, extracted landmarks that were used with an Extended Kalman Filter, and proposed a method that could be parameterized to trade-off exploring new areas with uncertainty. 
Martinez \etal~\cite{martinez2007active} reduced pose and map estimates with Gaussian Processes. 
The work of Rekleitis introduced an exploration versus exploitation framework to reduce uncertainty for a single robot by visiting previously mapped areas for single~\cite{rekleitis2012simultaneous} and multi-robot systems~\cite{rekleitis2013multi}.
Zhang \etal~\cite{zhang2014ear,zhang2015uncertainty} employed hybrid metric-topological maps to reduce uncertainty. All these early works considered only the 2D case.

More recent studies have expanded the problem from 2D to 3D, with the main platform considered being quadrotors, although a few studies utilizing manipulators also exist~\cite{lopez2019maintaining}. 
% These studies are more related to the objective of this paper, but in general they are not providing visibility distance constraints and unlike Aqua2 latteral motions and on the spot yaw rotations where considered. Other major differences will be highlighted.
The work of Forster \etal~\cite{forster2014appearance} provided a method to minimize uncertainty of a dense 3D reconstruction, but it was based on a direct method that has weak performance underwater, and mostly fly-over motions were performed without robust obstacle avoidance. Penin~\cite{penin2017vision} introduced a framework for producing trajectories taking into account the field of view limitations of the camera, but it was restricted to tracking only 4 points in close proximity to each other, no obstacles were considered, and no real-time performance.

The work of Spica \etal~\cite{spica2017coupling} combined visual servoing with Structure from Motion, but their primary focus was mapping and their method did not consider obstacles and operations in cluttered environments.
Constante \etal~\cite{costante2018exploiting} proposed a photometric method to drive the robot close to regions with rich texture, but as with Forster \etal~\cite{forster2014appearance}, direct methods do not perform well underwater and the motions were constrained to fly-overs and near-hovering.

Sheckells \etal~\cite{sheckells2016optimal} provided an optimal technique for visual servoing with no obstacle avoidance and only one visual objective was considered for the duration of the trajectory. 
% For our case, with a forward moving robot with a forward looking camera set, such as Aqua2, without the capability of lateral motions, in most cases more than one visual objective needs to be considered to keep track. 
Additionally, the work of Nageli \etal~\cite{nageli2017real,nageli2017real2} focuses on visual-objective tracking, rather than achieving a navigation goal with robust localization, and the potential field method applied for obstacle avoidance could result in a local minimum in cluttered environments. 

Other related studies considered only one visual objective ~\cite{potena2017effective,falanga2018pampc,yang2019optimized,potena2019joint,lee2020aggressive} or did not consider obstacle avoidance~\cite{potena2017effective,murali2019perception,costante2018exploiting,spasojevic2020perception, indelman2015planning}. 
It is worth noting that Spasojevic \etal \cite{spasojevic2020perception} were indeed able to track a set of landmarks, but with the constraint that they should always be tracked; in this work the robot needs to choose which objective(s) should be tracked from each position along the path.

Greef \etal~\cite{greeffperception} utilized a camera mounted on a gimbal. However, since that method was based on a teach and repeat approach, it is not applicable for  unexplored environments. 
Given the camera configuration and the kinematics of the Aqua2~\cite{dudek2007aqua}, \invis{the study above,} neither the method of Zhou \etal~\cite{zhou2020raptor}, nor of Murali \etal~\cite{murali2019perception} which allow lateral motions and free on-the-spot yaw rotations are suitable.
Some techniques~\cite{falanga2018pampc,lee2020aggressive} consider only one target, but they  resulted in low-level controllers.\invis{, while our goal is to keep the planning component high level enough in order to encourage extensibility, complex and sophisticated performance.} 
A very recent result by Zhang and Scaramuzza~\cite{zhang2020fisher} proposed a new topological model for map representation that could be used for guaranteeing uncertainty reduction in the entire map, but a computationally expensive offline computation on a known map is needed before planning, limiting the scope of online applications.  

In the underwater domain, in the context of coverage, Frolov~\etal~\cite{frolov2014can} proposed a motion planning framework for reducing map uncertainty by revisiting areas of high uncertainty, while Chaves~\etal~\cite{chaves2016opportunistic} utilized loop closures to reduce uncertainty.  Work by Karapetyan~\etal~\cite{KarapetyanOceans2020} used vision based navigation to perform coverage of a shipwreck with no state estimation. 

% \textcolor{red}{
Recent work from different groups has emphasized the potential of the Aqua2~\cite{dudek2007aqua} platform by providing effective real-time underwater navigation methods. Manderson \etal~\cite{manderson2018vision} provided a deep learning-based approach for collision avoidance by training upon the decisions of a human operator,  Hong \etal~\cite{hong2020semantically} utilized deep learning for classifying obstacles to static and dynamic on top of a potential field-based planner for obstacle avoidance, while Xanthidis \etal~\cite{xanthidis2020navigation}  produced a very capable model-based navigation framework using path-optimization. 
% }

% \textcolor{red}{
Finally, on the front of perception-aware underwater navigation, Manderson \etal~\cite{manderson2020vision} provided an extension to their previous work. Similarly to~\cite{manderson2018vision}, this deep-learning technique was based on fitting on data collected by a human operator controlling the robot. The robot was taught to stay close to corals, and avoid collisions with corals and rocks. Despite the effectiveness of this technique, the proposed solution is (a) unable to fully exploit the kinematic abilities of the robotic platform the way AquaNav does, because it does not consider roll motions and is limited to human intuition, (b) is naturally constrained to navigate only in similar environments (coral reefs), and (c) the motion commands follow a very reactive behavior and a short decision window that was compensated for by following predefined local goals. 
% }

% \textcolor{red}{
On the other hand, AquaVis produces locally near-optimal motions for avoiding the obstacles, with no reliance on a potentially error-prone human training process.  It also produces efficient trajectories for safe navigation in cluttered environments, similar to AquaNav. 
More importantly, since it operates on point-clouds, localization could be maintained with any kind of structures with rich texture, without the limitations dictated by a training dataset. 
Moreover, it is able to incorporate third-party object recognition modules for monitoring objects of interest, without the need of the time and resource intensive training on the motion planning module.% Finally, it is applicable for robots with arbitrary numbers and configurations of cameras.% to navigate and observe multiple visual objectives along the path.
% }

\section{Overview}
\label{prob}
\pagebudget{0.25}
%\vspace{-0.1in}
The goal of AquaVis is the safe navigation of an underwater robot, such as the Aqua2, moving freely through a cluttered three-dimensional environment\invis{, although there are no explicit assumptions}. Such navigation should be accomplished while maintaining visibility of sparse visual objectives along its path.
Specifically, the paths executed by the robot should reach a specified goal while keeping the path length short, avoid obstacles, and maximize the number of states along that path from which at least one visual objective is visible.

The proposed navigation framework, operating in unknown environments, selects from among the observed areas the ones that satisfy the visibility objectives and guides the trajectory towards the most appropriate ones. For example, in the presence of feature rich clusters, the trajectory is morphed to keep these clusters in the field of view. Barring any prior or sensed information about the environment, the main driver is motion towards the destination while minimizing the distance travelled. 
Although our primary focus lies on underwater robots, there are no explicit assumptions or limitations introduced that prohibit applications on other platforms and domains.

\subsection{AquaNav Overview}

\begin{figure}[t]
\centering 
{\includegraphics[width=0.8\columnwidth, trim=1cm 5.1cm 1.25cm 1.2cm, clip=true]{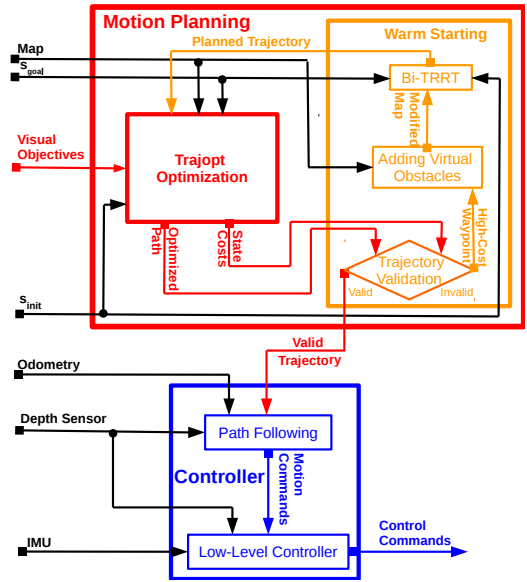}}
\caption{System architecture of AquaVis, which is based on AquaNav. AquaVis alters the core planning component by incorporating visual objectives, shown with red, while modules for warm-starting, shown with orange, and path following, shown with blue, are kept the same.} 
\label{fig:pipeline}
\end{figure}

To achieve the desired behavior of AquaVis, we utilized the robust navigation architecture of AquaNav~\cite{xanthidis2020navigation}, and extended its core planning module, utilizing Trajopt, to use the location of visual features as a constraint, Figure~\ref{fig:pipeline}.
In short, AquaNav is a waypoint navigation system, capable of real-time replanning and execution of trajectories with a guaranteed clearance, to ensure safety in the challenging and unpredictable underwater conditions. The AquaNav system is robust enough to enable real-time replanning, efficient and safe navigation in unknown environments, and is tested in real open-water conditions.
Trajopt is the primary path-optimization planner that ensures the above guarantees, and is assisted by a sampling-based warm-starting method, to overcome local minimum challenges.  This sort of path-optimization based approach not only generates high quality solutions rapidly, but also offers adequate flexibility, enabling modifications in the form of novel constraints and cost functions.
For the complete description of the AquaNav framework please refer to Xanthidis \etal~\cite{xanthidis2020navigation}.

\subsection{AquaVis Objective}

\begin{figure}[t]
\centering
\includegraphics[width=0.8\columnwidth, clip=true]{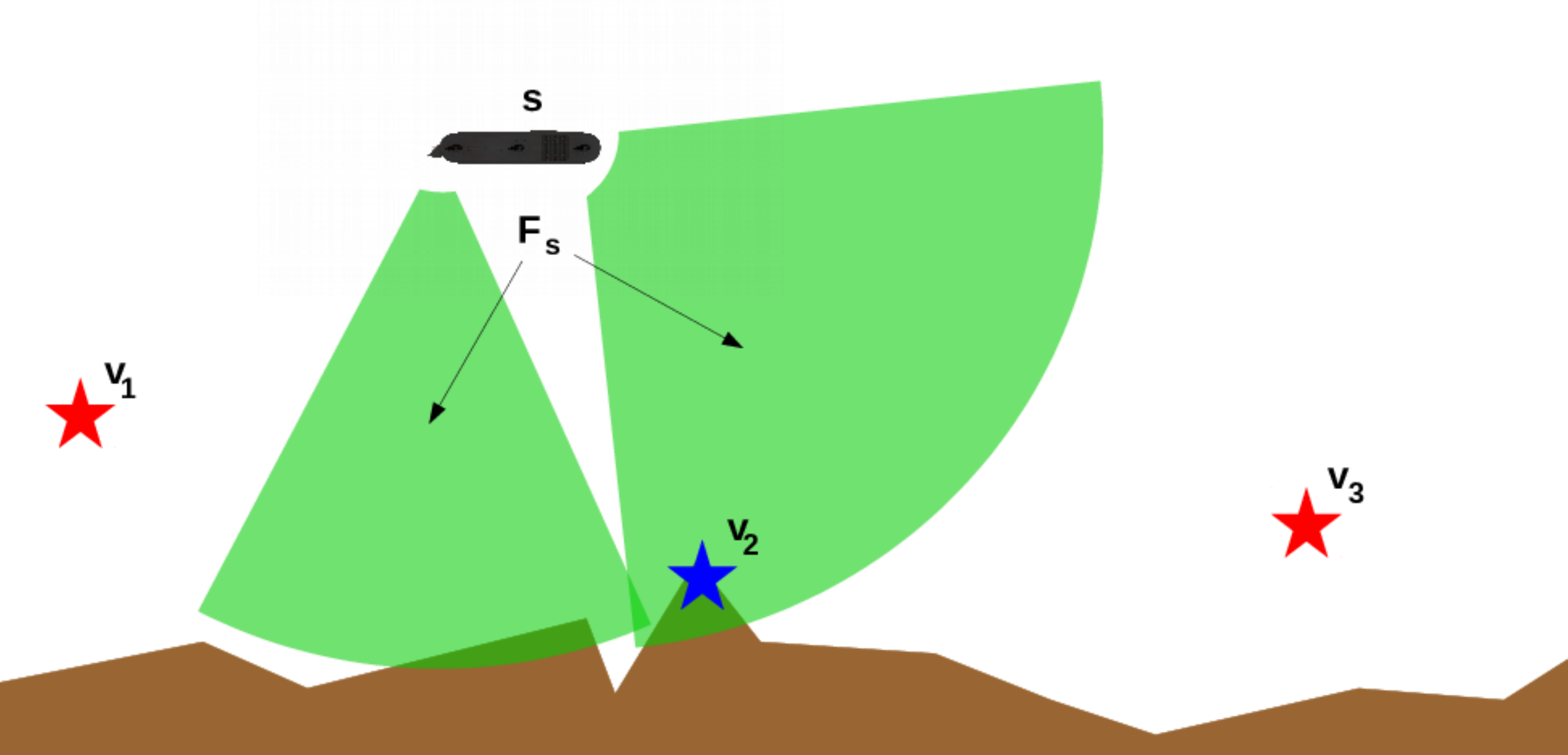}
\caption{Example of the visibility formulation used in Equation~\ref{eq_M_exec}. The visibility manifold $F_s$ for 2 cameras mounted on the robot is shown in light green. Visual objectives $v_1$, $v_2$, and $v_3$ are indicated with stars. Only $v_2$ is visible because it is inside $F_s$, while $v_1$ and $v_3$ are not observable from the robot's current state $s$.}
\label{fig:obj_yannis}
\end{figure}

% The robot's state $s \in SE(3)$ 
The robot's state $s$ 
describes its position and orientation in some fixed coordinate frame.  The robot is equipped with one or more cameras, such that from state $s$, a region $F_s \subseteq \mathbb{R}^3$ is visible from at least one of the cameras. Let $V$ denote a finite set of visual objectives.  For a given state $s$, each visual objective $v \in V$ may be visible (i.e. $v \in F_s$) or not ($v \notin F_s$); see~\fig{fig:obj_yannis} where visual objective $v_2$ is visible by the front camera.
% The path of the robot is represented as a sequence of waypoints $s_1,\ldots,s_n$.  
Additionally, let the continuous path of the robot be approximated by a sequence of consecutive states $s_1,\ldots,s_n$. 
Note that as $n$ increases, we can approximate the robot's continuous path with arbitrary precision.  
We quantify the path's success in maintaining visibility of the visual objects via the following function:
\begin{equation}
    M(s_1,\ldots,s_n) = \frac{\left| \left\{ s_i \mid F_{s_i} \cap V \ne \emptyset  \right\}  \right|}{n}
    \label{eq_M_exec}
\end{equation}
This function provides the fraction of the states in the path that observe at least one objective. It reaches $1$ if all the states are able to observe at least one visual objective, or $0$ if no visual objectives were observed during traversing the entire path.
Thus, the objective of AquaVis is to minimize the path length,  avoid obstacles, and maximize Equation~\ref{eq_M_exec}.

\section{Proposed Approach}
\label{prop_app}
%\vspace{-0.1in}
\pagebudget{1.5}
This section describes the enhancements of AquaVis upon the AquaNav pipeline. These enhancements consist of ways to automatically extract the visual objectives, modify the planning process to accommodate them, and ensure the satisfaction of the kinematic constraints. 

\subsection{Extracting Visual Objectives}
With respect to the AquaVis pipeline (Figure~\ref{fig:pipeline}), the visual objectives are considered as an input in the form of a list of 3D points.
These visual objectives are either user-defined, or automatically extracted online. For example, known methods that detect corals~\cite{modasshir2018mdnet,modasshir2018coral,modasshir2019autonomous}, or other Aqua2 robots~\cite{joshi2020deepurl} and extract the 3D positions of those features could be employed for application specific purposes, such as environmental monitoring or multi-robot exploration.

% The focus of this paper lies on improving odometry estimates, instead of tracking objects of interest. 
Visual objectives could be extracted automatically to assist underwater state estimation, by utilizing the output of most SLAM techniques.
In particular, AquaNav employs the robot to navigate through an unknown environment, using a  state estimation package, such as SVin2~\cite{RahmanIROS2019a}, that outputs both the odometry and a representation of the sensed environment as a 3D point cloud.
Thus, the raw point-cloud could be processed to extract visual objectives with high density of features, then these visual objectives could assist the odometry as landmarks\invis{ assuming that the camera is able to track them}. 
Such an approach is a necessity in the underwater domain, which is notoriously challenging for vision-based state estimation~\cite{QuattriniLiISERVO2016,joshi2019experimental}, in part because the quality of the features is often low, and their spatial distribution uneven, with most features concentrated in only a few places.

We propose extracting visual objectives from a point-cloud by treating the problem as density-based clustering. DBSCAN~\cite{ester1996density} is applied on the 3D point cloud to detect clusters with high density and then the centroids of these clusters are chosen as the visual objectives. 
DBSCAN has a minimal number of parameters: the minimum number of samples per cluster and the minimum proximity. The operator decides the quality of the objectives to be tracked, in terms of both number and density of good features~\cite{ShkurtiCRV2011}.

In each planning cycle, the above preprocessing step produces the visual objectives used during planning. Though, not keeping past information of previously detected clusters, could result to a highly sub-optimal reactive behavior. Thus, the set $V$ contains a maximum of $m$ computed visual objectives, in order to ensure real-time planning, and to avoid excessive computation from an ever increasing number of visual objectives. Initially, the visual objectives are added to the list until $|V| = m$. Then, any new measurement replaces the closest one if they are in close proximity by updating the center of the cluster, or in any other case, the oldest one to favor locality and computational efficiency.

\subsection{Motion Planning Modifications}
AquaVis modifies the path optimization element of the AquaNav framework, which is built upon the optimization-based package Trajopt. A brief review of the original Trajopt formulation is discussed next.

\subsubsection{Original Trajopt formulation}
Trajopt attempts to minimize the function
\begin{equation}
    f(S) = \displaystyle\min_{S}
    \sum\limits_{i=1}^{n-1}||s_{i+1}-s_i ||,
    \label{eq:trajopt_obj}
\end{equation}
where $S = \langle s_0, s_1, \dots, s_n \rangle$ the sequence of $n$ states of the robot considered during optimization. $f(S)$ is the sum of squared displacements, which minimizes  path length.

Collision constraints are enforced for every state $s \in S$:
\begin{equation}
    h(s) = \sum\limits_{o\in O}|d_{\rm safe} - \operatorname{sd}(PC_s,o) |
    \label{eq:h_s}
\end{equation}
where $O$ is the set of obstacles, $PC_s$ is the 3D geometry of the robot in state $s$, and $\operatorname{sd}$ represents the minimum Euclidean distance to separate two 3D convex objects. More details about $\operatorname{sd}$ appear in~\cite{schulman2014motion}.

The above constraint, given successful convergence, guarantees that each waypoint on the path will maintain distance at least $d_{\rm safe}$ from the closest obstacle, but has no guarantees on the transitions between waypoints. To enforce continuous time safety, instead of Equation~\ref{eq:h_s}, the following function is applied for each pair $s_{i-1}, s_{i}$ of consecutive states:
\begin{equation}
    H(s_i, s_{i+1}) = \sum\limits_{o\in O}\left|d_{\rm safe} - \operatorname{sd}(L^{s_i}_{s_{i-1}},o) \right|,
    \label{eq:HH_s}
\end{equation}
in which $L^{s_i}_{s_{i-1}}$ is the convex hull of $PC_{s_{i-1}} \cup PC_{s_{i}}$. 

\begin{figure}[t]
 \begin{center}
  \leavevmode
   \begin{tabular}{cc}
     \subfigure[]{\includegraphics[width=0.4\columnwidth, clip=true, trim={1.5in 1.5in 1.5in 0.4in}]{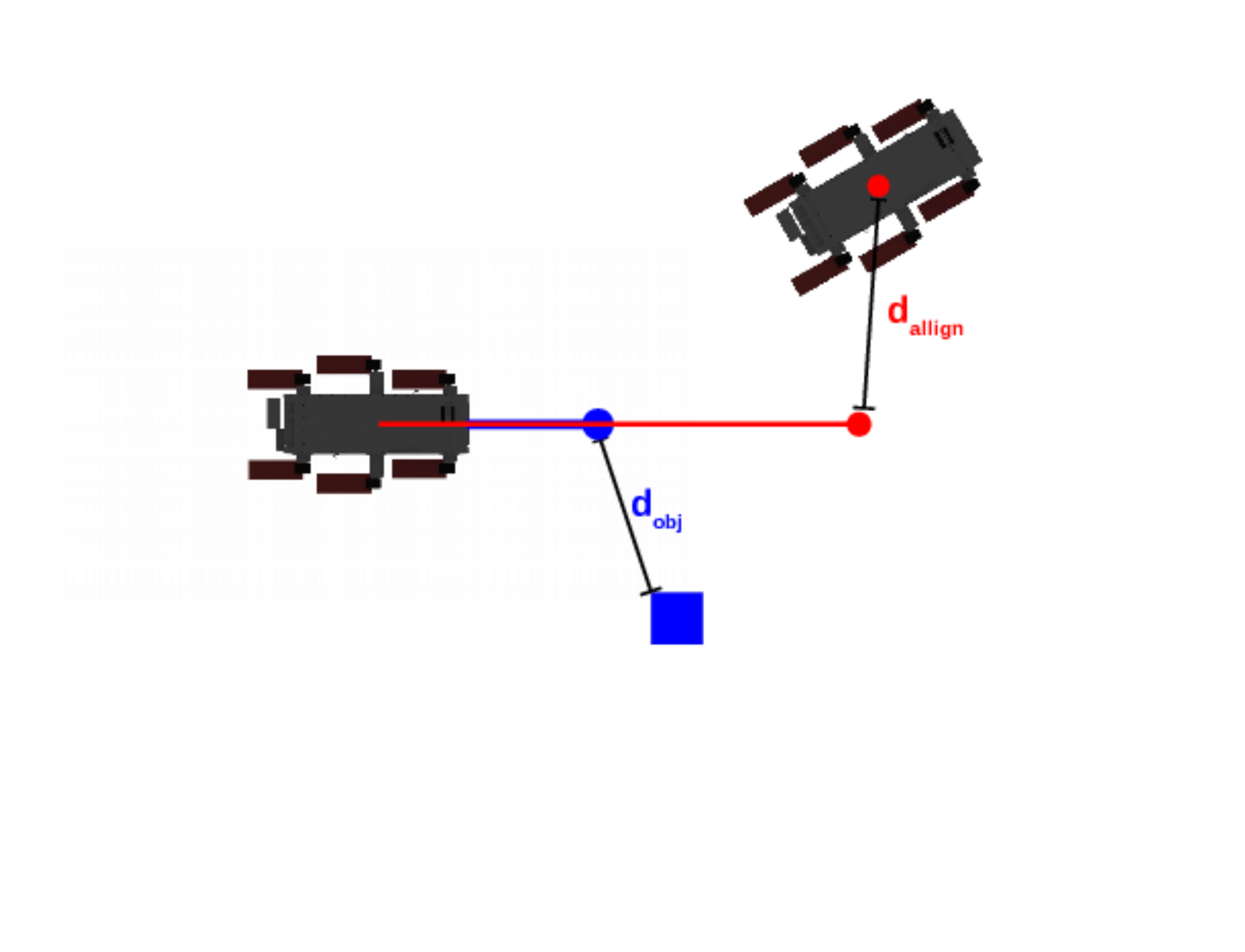}}&
     {\subfigure[]{\includegraphics[width=0.4\columnwidth, clip=true, trim={1.1in 0.8in 1.1in 1.1in}]{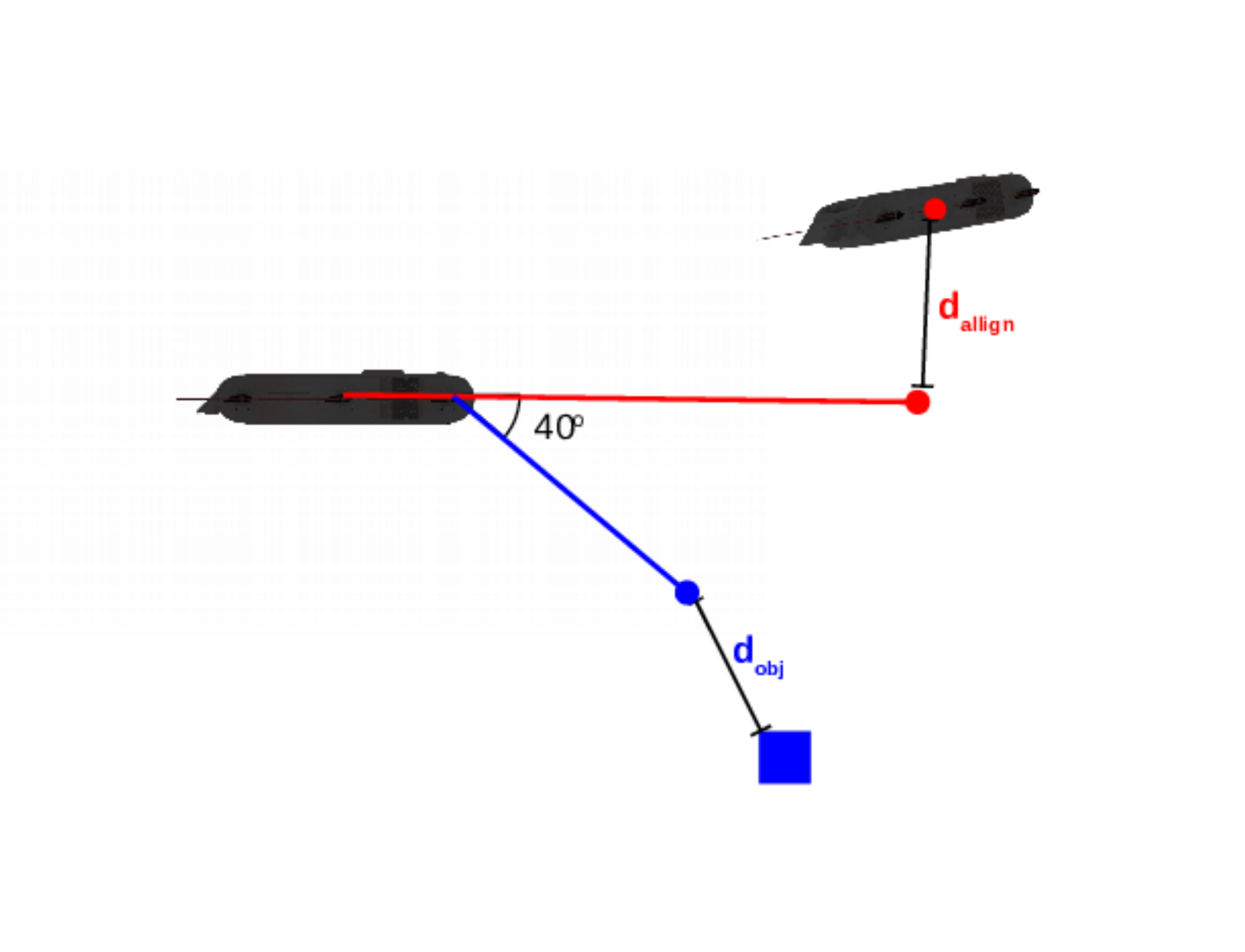}}}
    \end{tabular}
  \end{center}
\caption{Top (a) and side (b) views of a state using the novel constraints during optimization. The blue square indicates a visual objective, and the red circle marks the next waypoint. Minimizing $d_{obj}$ will result on the robot observing the objective, while minimizing $d_{align}$ will result on the robot to be consistent with the kinematics assumed during path execution and planning.}
\label{fig:constraints}
\end{figure}

Two additional cost functions are introduced. 
The first new cost function incentivizes the robot to view visual objectives, whereas the second one forces the path to self-correct and maintain the kinematic constraints to visit each waypoint assumed by the path follower.   
The functions are described below, and an example that outlines these novel cost functions is shown in Figure~\ref{fig:constraints}. 

\subsubsection{Visibility Constraints}

\begin{figure}[t]
 \begin{center}
  \leavevmode
   \begin{tabular}{cc}
     \subfigure[]{\includegraphics[width=0.4\columnwidth, clip=true, trim={1.0in 0.0in 1.0in 0.0in}]{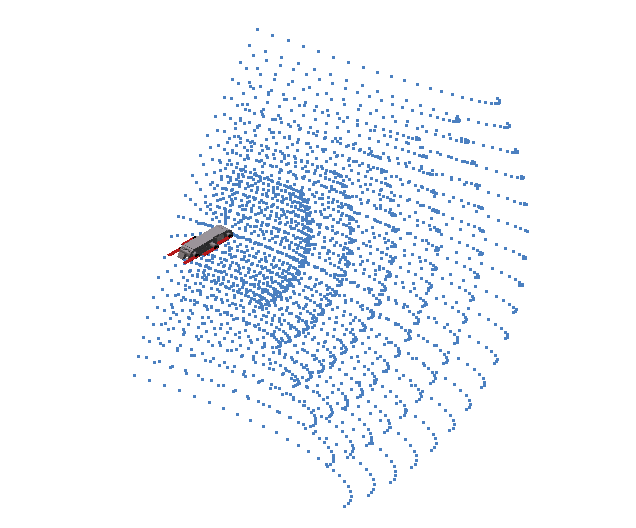}}&
     \subfigure[]{\includegraphics[width=0.4\columnwidth, clip=true, trim={1.0in 0.0in 0.5in 0.0in}]{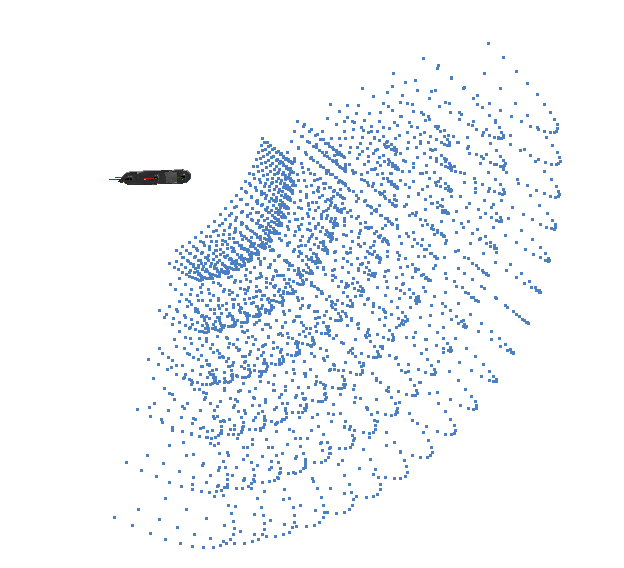}}
    \end{tabular}
  \end{center}
\caption{Different perspectives of the projected points of the $F_s^{\sim}$ visibility set approximating the $F_s$ visibility manifold corresponding to the front camera.}
\label{fig:points}
\end{figure}
The visibility constraint is intended to direct the robot to observe a known set of visual objectives. 
The core idea is to project a set of points $F_s^{\sim}$ in front of the robot's cameras to approximate $F_s$ and then attempt to minimize the distance $d_{obj}$ between the closest visual objective to the closest projected point of $F_s^{\sim}$. By minimizing the above distance to zero, the robot is guaranteed to track at least one visual objective in state $s$. Figure~\ref{fig:points} shows an example of the above concept for the front camera of the robot.

% Two items in the sequence == no comma
% Apples and bananas
% Apples, bananas, and grapes
%
Given a  set of objectives $V$ and a  visibility set $F_s^{\sim}$ for the state $s$, the general form of the proposed constraint applied to each state is:
\begin{equation}
    \operatorname{Vis}(s) = \smash{\displaystyle
    \min_{v \in V}
    \min_{f \in F_s^{\sim}}
    \left|\left|f-v \right|\right|
    }
    \label{eq:gen_vis}
\end{equation}
Trajopt was modified to utilize the above cost function to minimize the distance $d_{obj}$ between the projected desired point and the nearest visual objective. 
So, given successful convergence, at least one visual objective will be visible at a desired direction and distance for each state. 

It should be noted that there is an important trade-off between approximating $F_s$ accurately and real-time performance. 
During the optimization process, for each state, the distance between each visual objective and each projected point $f$, $f \in F_s^{\sim}$ is calculated, resulting potentially to slow re-planning. Real-time performance requires selecting a small set $F_s^{\sim}$, thus further relaxing path optimality. 
% Especially, even very limited $F_s^{\sim}$ sets, can result into an acceptable performance, by trading path optimality for computational efficiency.
\begin{figure*}[h]
\leavevmode
\begin{center}
   \begin{tabular}{ccc}
     \subfigure[]{\includegraphics[height=0.12\textheight, clip=true, trim={1.0in 3.0in 3.0in 0.0in}]{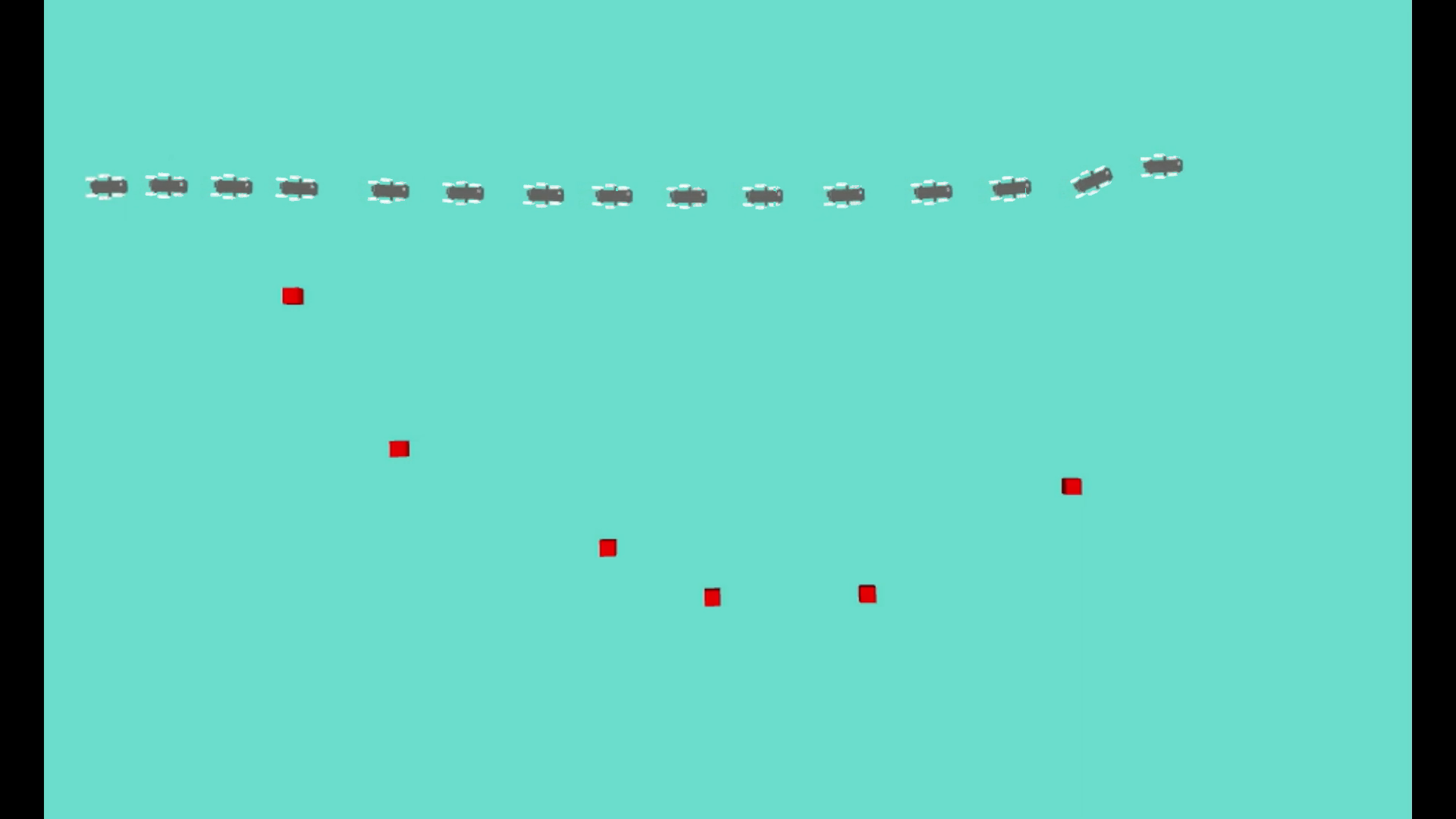}}&
     \subfigure[]{\includegraphics[height=0.12\textheight, clip=true, trim={1.0in 3.0in 3.0in 0.0in}]{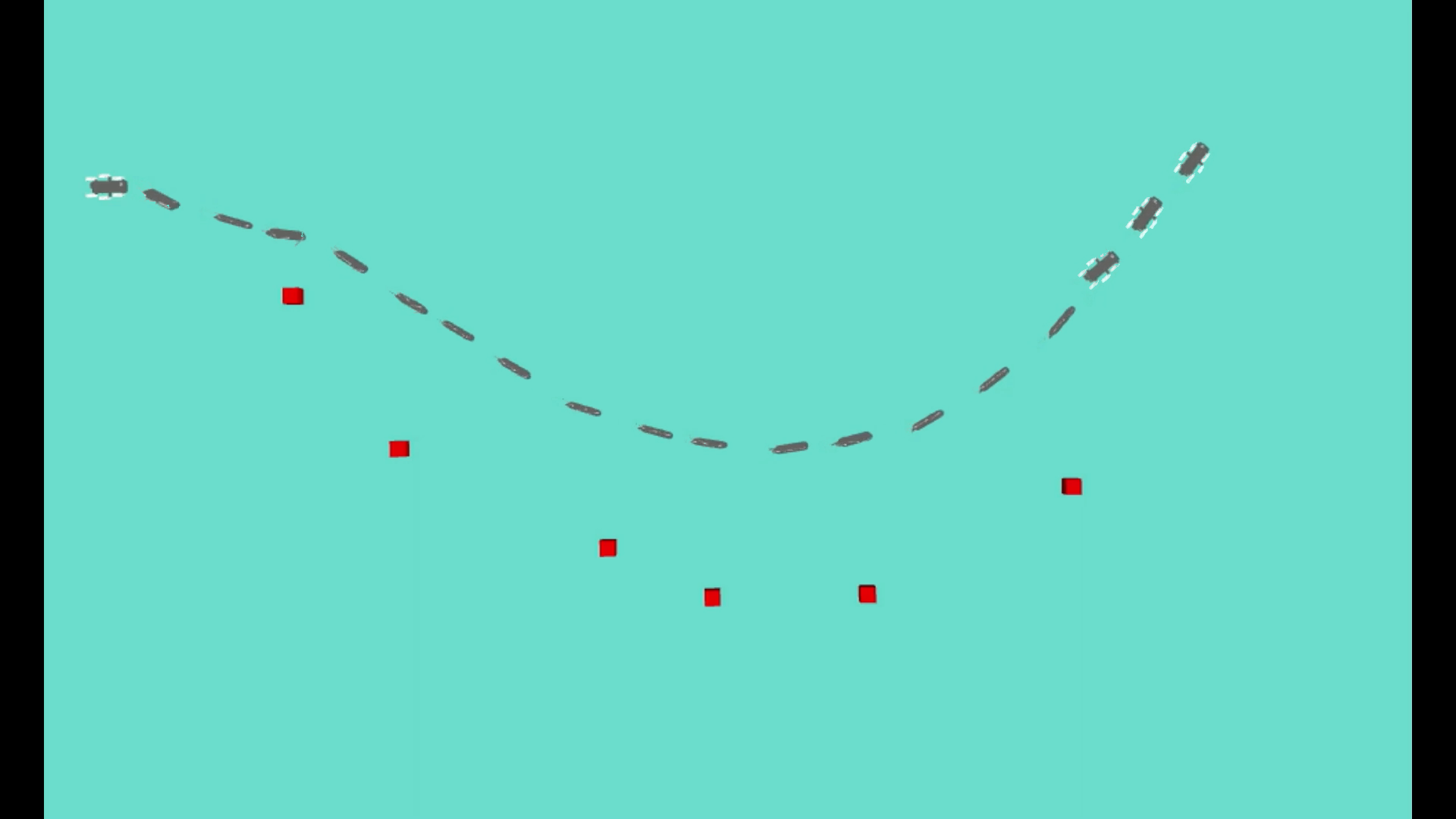}}&
     \subfigure[]{\includegraphics[height=0.12\textheight, clip=true, trim={0.0in 0.0in 0.0in 0.0in}]{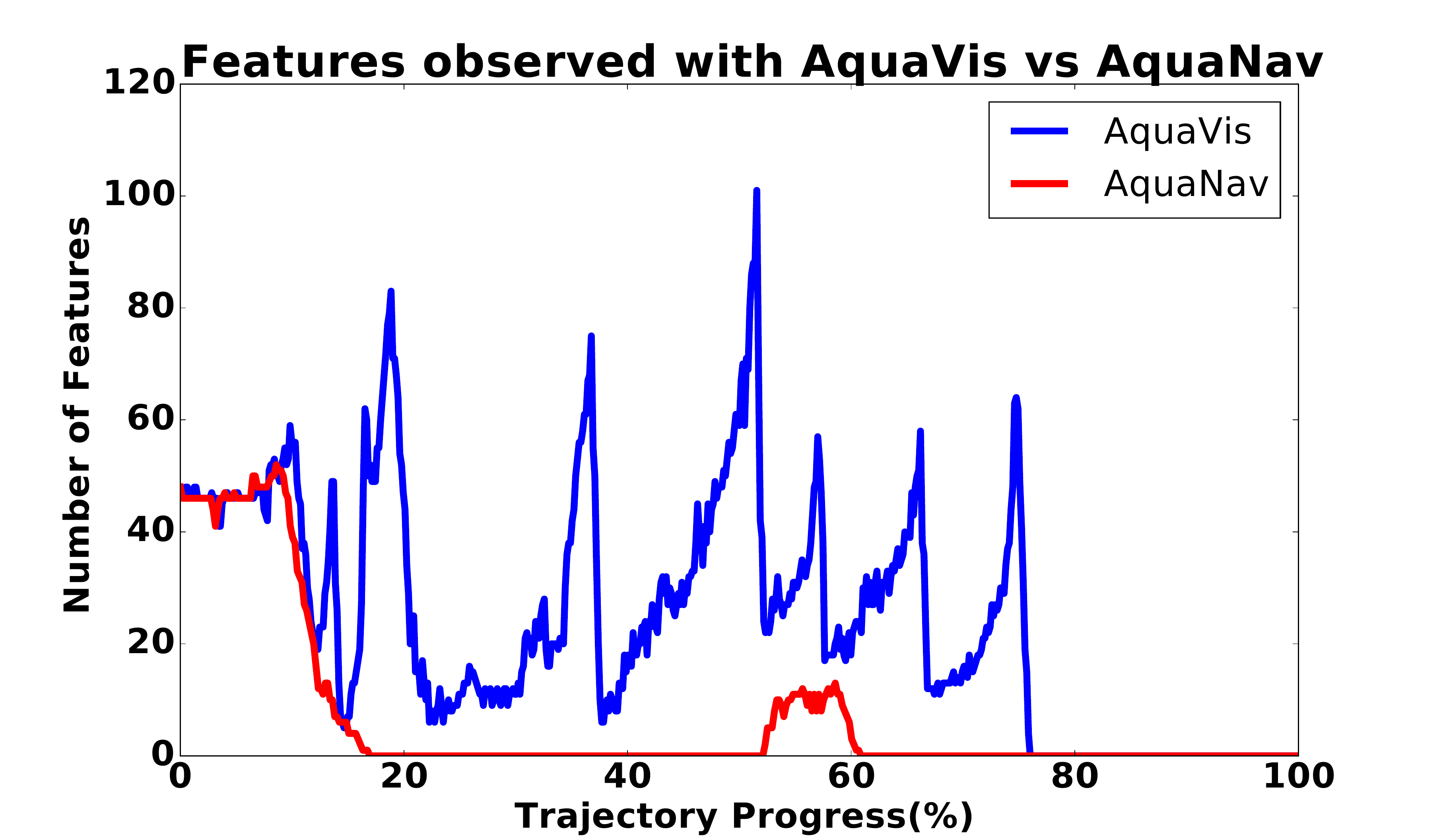}} \\
     \subfigure[]{\includegraphics[height=0.12\textheight, clip=true, trim={1.5in 3.0in 3.0in 0.0in}]{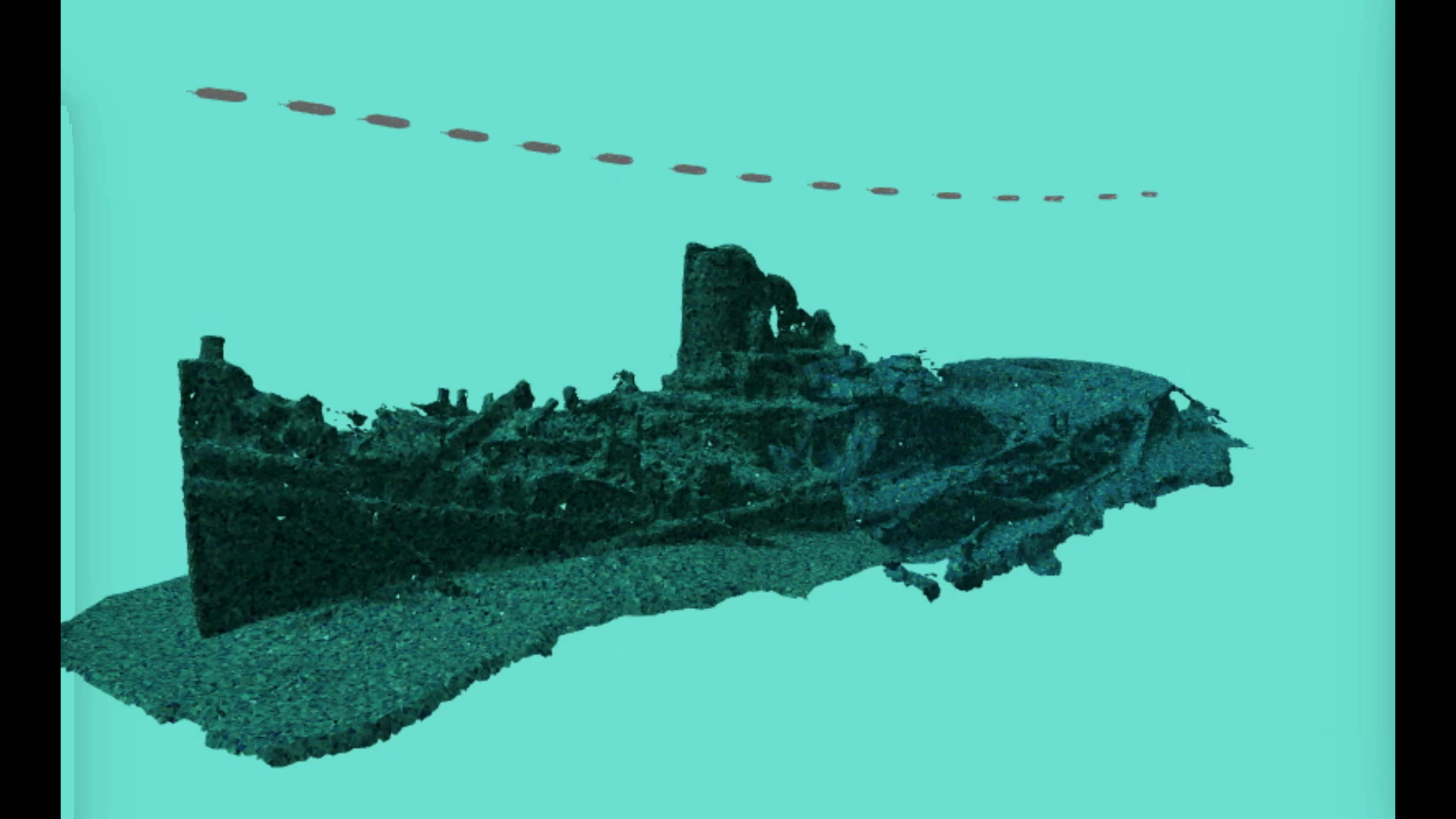}}&
     \subfigure[]{\includegraphics[height=0.12\textheight, clip=true, trim={1.5in 3.0in 3.0in 0.0in}]{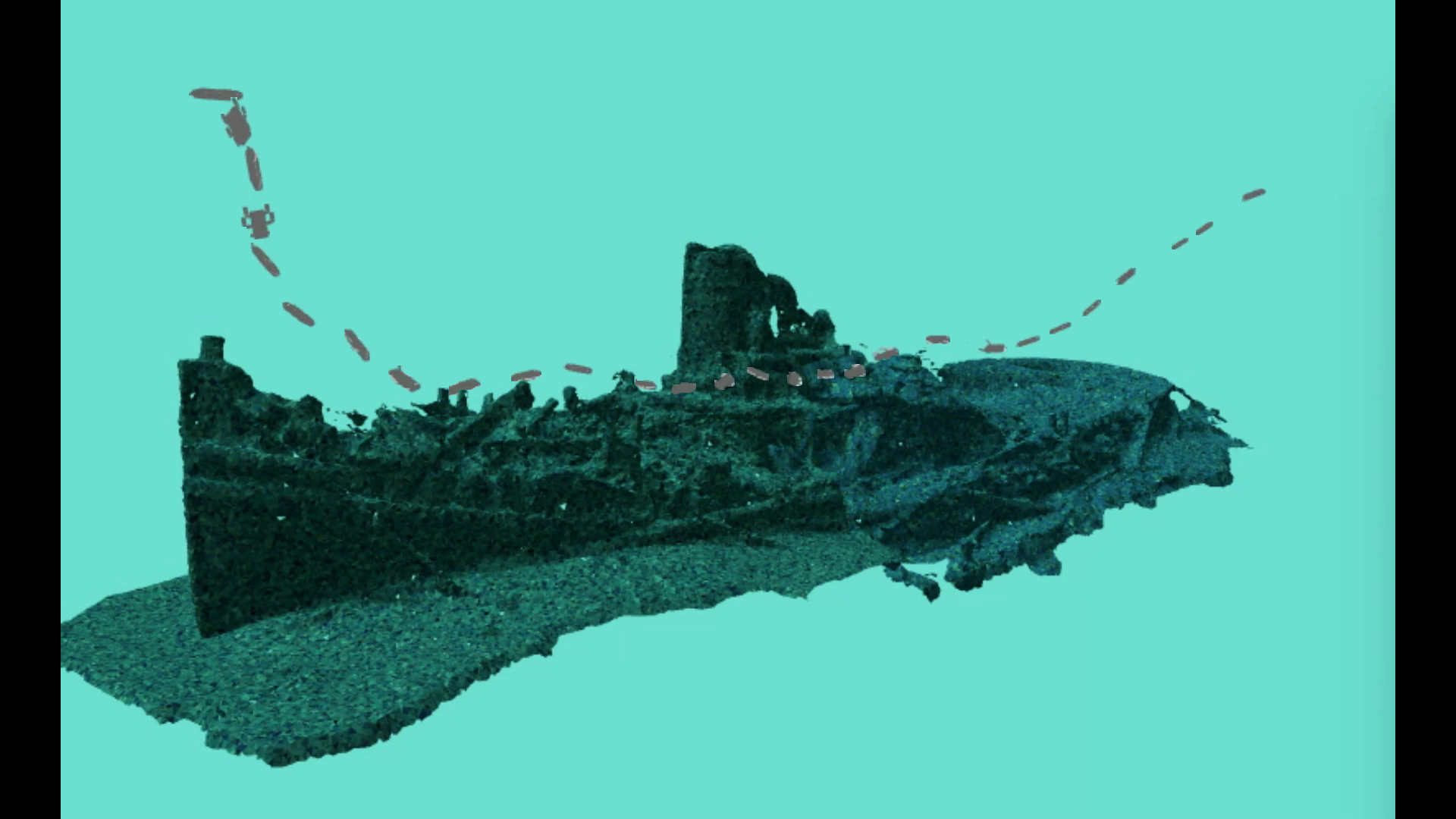}}&
     \subfigure[]{\includegraphics[height=0.12\textheight, clip=true, trim={0.0in 0.0in 0.0in 0.0in}]{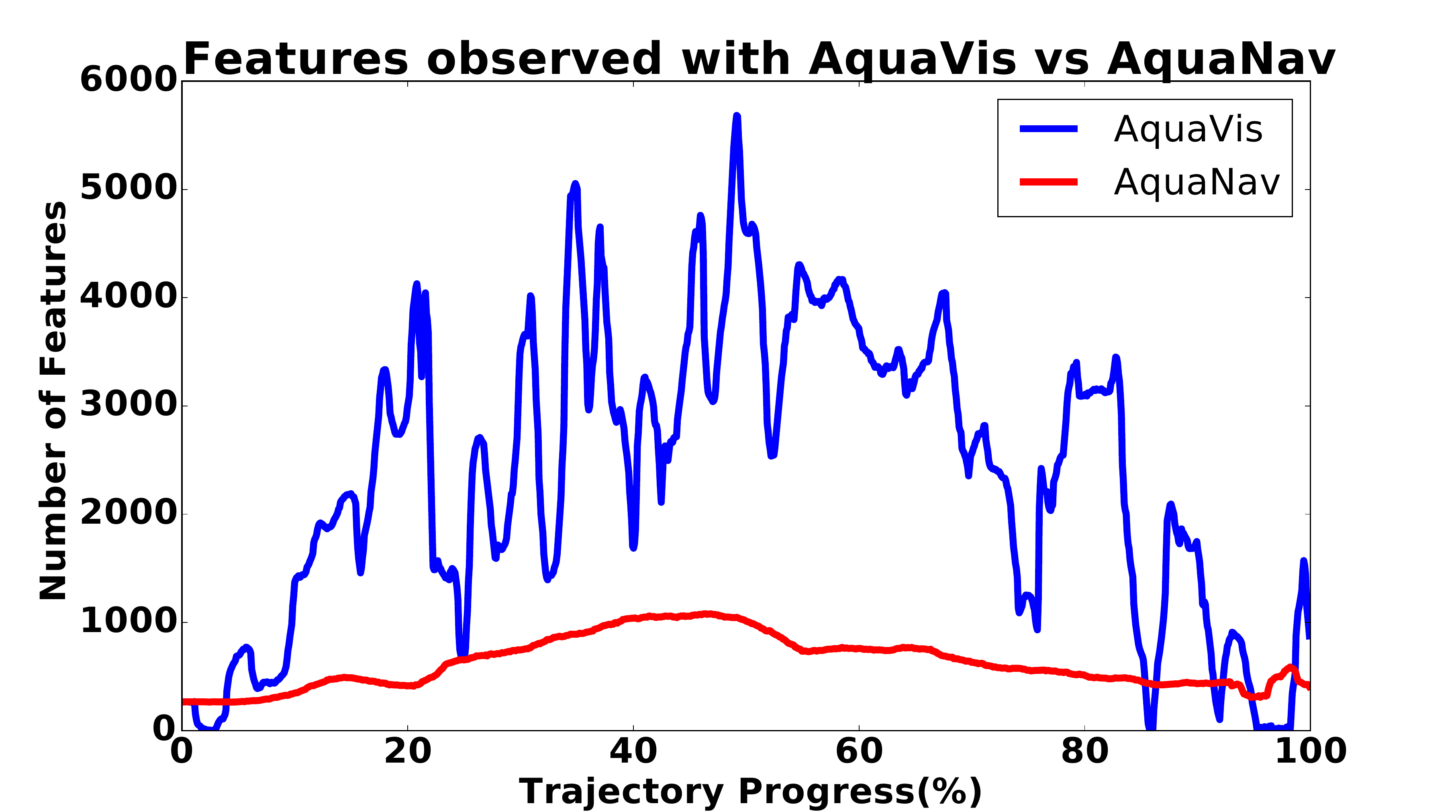}}
    \end{tabular}
\caption{The results for the Pilars environment are shown on the top, and for the Shipwreck at the bottom row. The trajectories produced by AquaNav are shown in (a) and (d), and for AquaVis at (b) and (e). The features observed for both methods are shown in (c) and (f).}
\label{fig:exp}
\end{center}
\end{figure*}

Acceptable performance that trades path optimality and smooth transitions for low computational cost, and fast re-planning, could even be achieved with a set $F_s^{\sim}$ formed by a single point in front of the camera at a desired distance. 
Let $T_w^s$ be the transformation from the world reference frame to the robot's local reference frame at state $s$, $T_r^c$ is the transformation from the robot's local reference frame at state $s$ to the camera $c$, $C$ is the set of cameras, and $d_{\rm vis}$ is the desired distance of observing a visual objective, then:
% \begin{equation}
%     F_s^{\sim} = \{T_w^s T_r^c \begin{bsmallmatrix} d_{\rm vis} \\ 0 \\ 0 \\ 1 \end{bsmallmatrix} \mid c \in C\}
%     \label{eq:spec_vis}
% \end{equation}
% % \mid instead of | gives a little better spacing
\begin{equation}
    F_s^{\sim} = \bigcup_{c \in C} \left\{T_w^s T_r^c \begin{bsmallmatrix} d_{\rm vis} & 0 & 0 & 1 \end{bsmallmatrix}^T \right\}
    \label{eq:spec_vis}
\end{equation}
This formulation guarantees that during the planned trajectory, multiple objectives could be observed from multiple cameras, contrary to the works discussed previously. Also with the simplification described in Equation~\ref{eq:spec_vis}, AquaVis achieves real-time behavior with no significant added delays to the AquaNav replanning baseline. Moreover, although path-optimality was sacrificed for real-time performance, in our experiments it is shown that efficient and smooth paths were produced thanks to the greedy nature of the formulation combined with fast replanning.

\subsubsection{Kinematic Constraints}
\invis{
In the previous section, a cost function steering the robot towards observing visibility objectives was proposed. It should be expected that although the states might produce a trajectory where such visibility objectives are achieved, the transitions might include motions that cannot be executed by a non-holonomic robot, such as Aqua2. Thus, a second constraint must be introduced to guarantee that the resulting trajectories not only satisfy the objective, but also obey the kinematics and can be executed by the robot.}
The only objective of AquaNav was to minimize the path-length in terms of both translation and rotation. So the produced plans could be executed directly by a way-point follower, since unnecessary aggressive rotations are not expected.
Similarly, the paths produced by using only the visibility constraints could be directly executed by a holonomic robot, enabling it to move and rotate in way to track the necessary visual objectives. 
However, the Aqua2 vehicle is not a holonomic robot, thus by using only the visual constraints, many motions requiring lateral translation could not be executed by the path follower.
To ensure that the resulting trajectory for observing the visibility objectives satisfies the kinematic constraints of the vehicle a second constraint is introduced. 
The path follower module in the AquaNav pipeline accepts the 3D coordinates that need to be reached by the robot, along with a constant desired roll orientation during the motion. 
Ideally, the robot would move along the straight line segments connecting successive waypoints. 
Thus, the robot after achieving the waypoint $p_{s_i}$ should maintain an orientation pointing directly to the next waypoint $p_{s_{i+1}}$, with $p_{s_i}$ indicating the 3D coordinates of state $s_i$.

The cost function utilized by AquaNav for aligning the robot properly during planning is similar to the cost function described for the visibility constraints: A point is projected in front of the robot at a specific distance and the distance $d_{\rm align}$ of this point to the next waypoint is minimized.
More precisely, let $S= [ s_1, s_2, \dots, s_{n-1}, s_{n}]$ be the trajectory to be optimized, the cost function applied for each state $s_i$ is
% \begin{equation}
% A(s_i) = 
% \begin{cases}
%     \left|\left|T_w^{s_{init}}\begin{bsmallmatrix} 
%     av(S)- \epsilon
%      \\ 0 \\ 0 \\ 1 \end{bsmallmatrix} -\begin{bsmallmatrix} p_{s_{1}}^T \\ 1 \end{bsmallmatrix}\right|\right| , & \text{if } i=1\\ \text{ } \\
%     \left|\left|
%     T_w^{s_i}\begin{bsmallmatrix} av(S)- \epsilon \\ 0 \\ 0 \\ 1 \end{bsmallmatrix} -\begin{bsmallmatrix} p_{s_{i+1}}^T \\ 1 \end{bsmallmatrix}
%     \right|\right|,              & \text{otherwise,}
% \end{cases}
% \label{eq:spec_align}
% \end{equation}
\begin{equation}
A(s_i) = 
    \left|\left|T_w^{s_{i}}\begin{bsmallmatrix} 
    av(S)- \epsilon
     \\ 0 \\ 0 \\ 1 \end{bsmallmatrix} -\begin{bsmallmatrix} p_{s_{i+1}}^T \\ 1 \end{bsmallmatrix}\right|\right|,
\label{eq:spec_align}
\end{equation}
where $p_{s_{i}}$ is the coordinates of state $s_i$ in the world frame, and $av(S)$ is the average length of the $S$ trajectory given by:
\begin{equation}
    av(S) = \frac{\sum\limits^{n}_{i = 2}\left|\left| p_{s_{i}}^T   -  p_{s_{i-1}}^T \right|\right|}{n-1}
    \label{eq:ave_length}
\end{equation}
The first element of the first vector, similar to Equation~\ref{eq:spec_vis}, is the distance the point will be projected, in this case forward. 
The distance is calculated as the average distance between two consecutive states reduced by a small positive value $\epsilon$. 
So the projected point is adjusted automatically to be the same for every state to encourage consistency, while the $\epsilon$ factor is used to encourage shorter path lengths.
\invis{
One limitation is that, within the current formulation, tracking specific visual objectives is not guaranteed on the transitions between the different waypoints. But our expectation is that such short-term issues could be mitigated with constant replanning, and in the worst-case, with the loop closure capabilities of many vision-based SLAM frameworks, such as SVIn.
}

\section{Experimental Results}
\label{exper}
%\vspace{-0.1in}
\pagebudget{1.75}
The performance of AquaVis was validated in simulation. Its flexibility in controlling the trade-off between path-length, tracking visual objectives, and satisfying the way-point navigation kinematics was explored. 
In our experiments the desired clearance for obstacle avoidance, similarly to \cite{xanthidis2020navigation}, was set to \SI{0.6}{m}, the desired visibility distance $d_{\rm vis}$ to \SI{1.0}{m}, and the linear velocity of the robot to \SI{0.4}{m/s}.

Additionally, the original camera configuration of Aqua2 --- our target system --- was used. Aqua2 leverages a forward-looking stereo camera system for state estimation. That system has a field of view \ang{90} vertically and \ang{120} horizontally.  The forward-looking camera is tilted downwards by \ang{40} to further assist state estimation.

%\vspace{-0.1in}
\subsection{Simulations}

We simulate the detected features from the stereo VIO with a lidar sensor that returns 3D clusters of features in select obstacles. 
In the real system, fewer features will be detected but that does not affect the planning process negatively, instead the process of extracting visual objectives is expected to be executed faster. 
The simulated lidar has the same field of view with the Aqua2 front cameras, a resolution of $100\times75$, and range of \SI{6}{m}, to represent the expected turbidity of the underwater domain. 
To extract visual objectives automatically using DBSCAN~\cite{ester1996density}, the maximum distance between features was set to \SI{0.2}{m} with a minimum number of $5$ features per cluster. A maximum set of 15 visual objectives was maintained, with new visual objectives replacing the closest of the old ones that were in a distance less than \SI{0.5}{m}, or the oldest in the set.
AquaVis is tested online against AquaNav in 2 different environments, the Pilars, and the Shipwreck shown in Figure~\ref{fig:exp}.

The Pilars Environment, shown in Figure~\ref{fig:exp}(a-c), is intended to test AquaVis in an environment where feature rich areas are distributed sparsely in the environment; a top down view is presented. \invis{which is a common scenario in the underwater domain}
AquaNav, by optimizing path length, moves on a straight line, disregarding  the features, which are essential for localization. 
AquaViz, in contrast, reaches the same goal while passing in proximity and observing the feature rich areas (red cubes). 
The plot in Figure~\ref{fig:exp}(c) confirms our expectations: AquaNav cannot observe any features for the majority of the time, whereas AquaVis consistently tracked enough features. It is worth noting, that AquaVis introduced a $90^\circ$ roll to bring the visual objectives of the pilars in the field of view.  Moreover, AquaVis maintained tracking for the first $75\%$ of the trajectory that visual objectives could be observed, and lost track at the last $25\%$ where no visual objectives were present that could be observed with a forward looking camera.

Similarly, in the Shipwreck environment, shown in Figure~\ref{fig:exp}-(d-f), AquaVis was able to observe consistently more features than AquaNav, excluding ascent and descent, which is expected given the kinematics of Aqua2.
Also the robot not only oriented itself to track most of the shipwreck but also created the desired proximity, indicating potential use for mapping purposes. 
On the other hand, AquaNav moved in a straight line, unaware of the feature rich areas, and tracked only a small portion of the features tracked by AquaVis.
\begin{figure}[t]
 \begin{center}
  \leavevmode
   \begin{tabular}{cc}
     \fbox{\subfigure[]{\includegraphics[width=0.4\columnwidth]{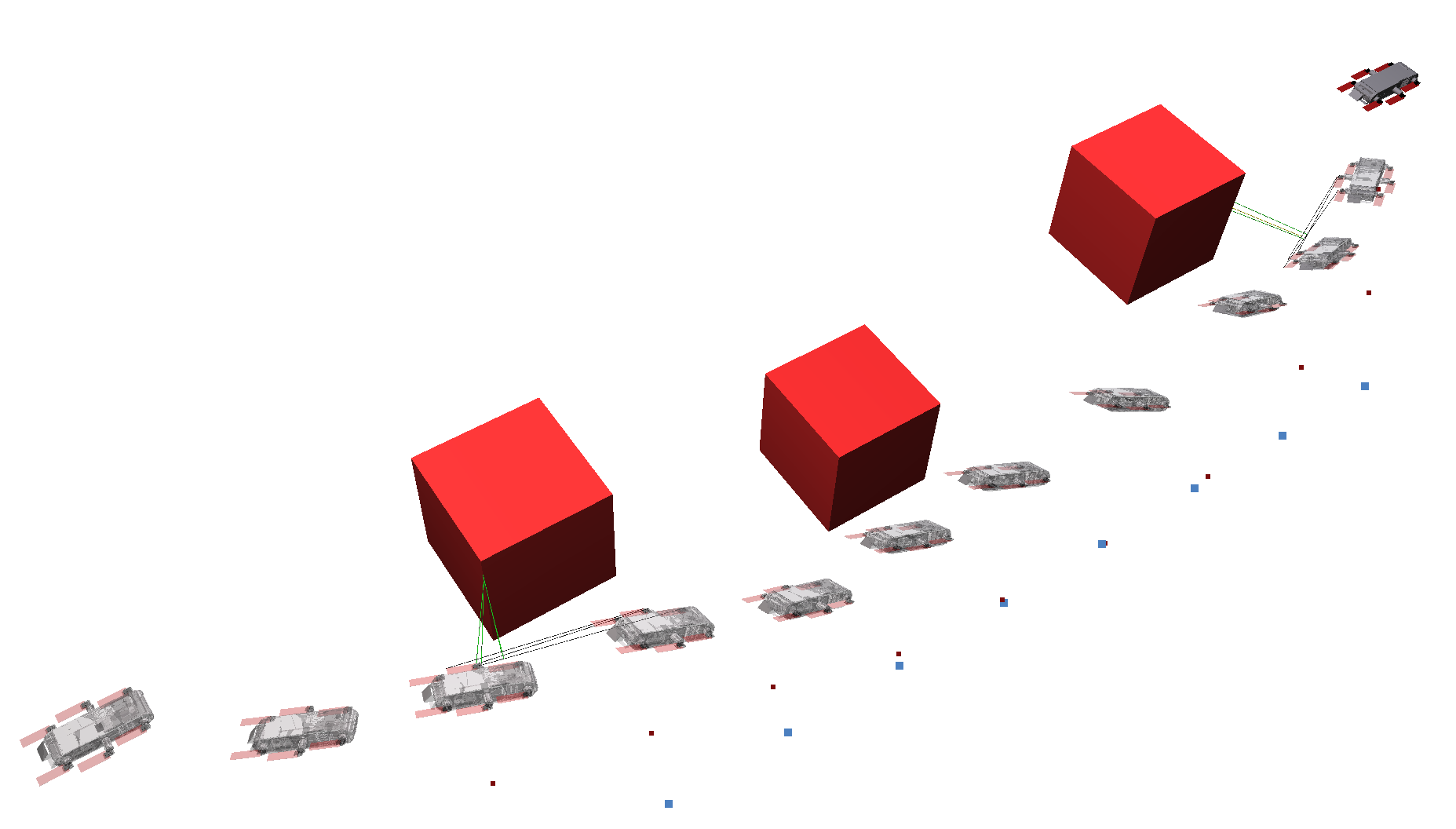}}}&
     \fbox{\subfigure[]{\includegraphics[width=0.4\columnwidth]{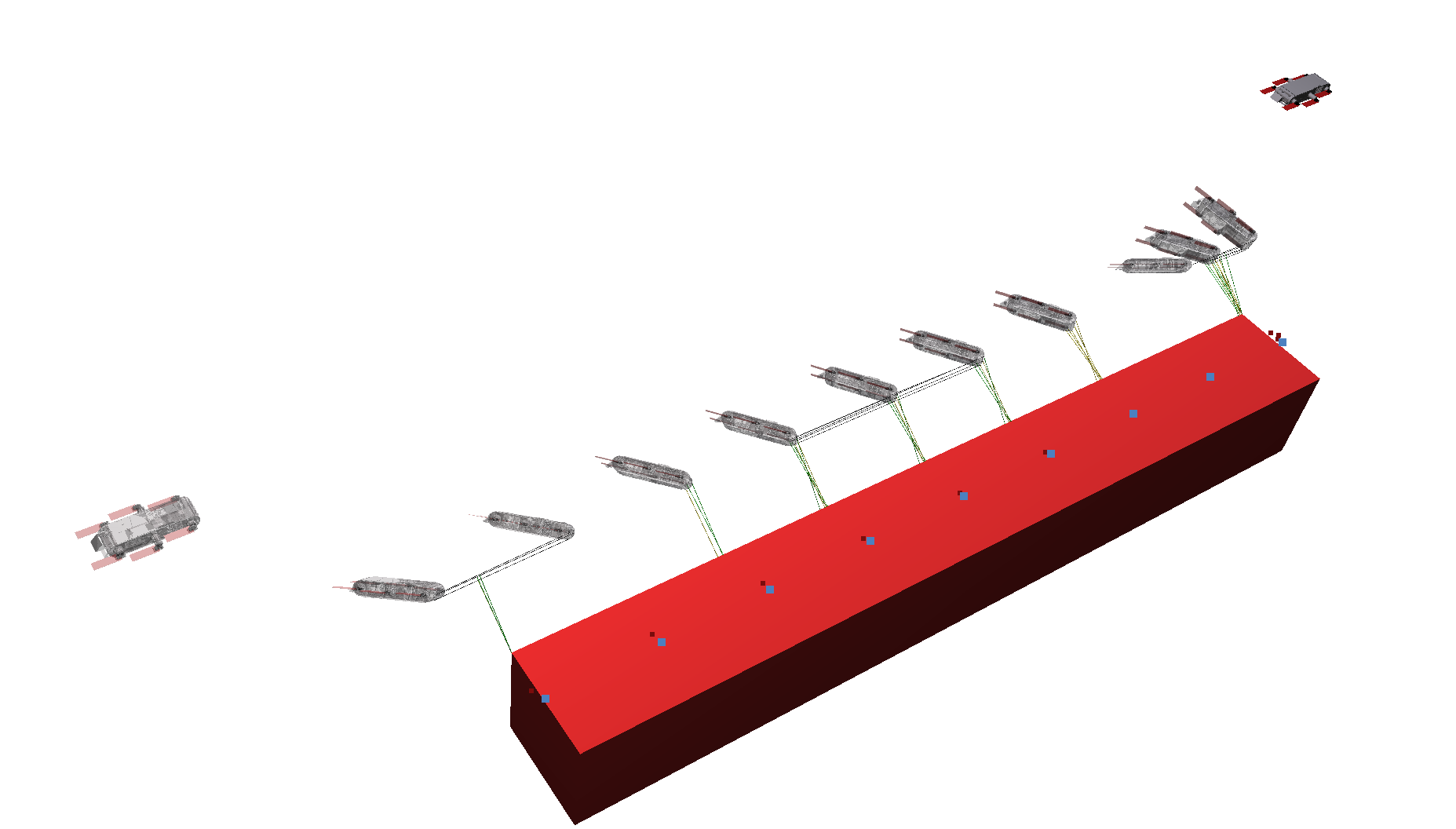}}}\\
     \fbox{\subfigure[]{\includegraphics[width=0.4\columnwidth]{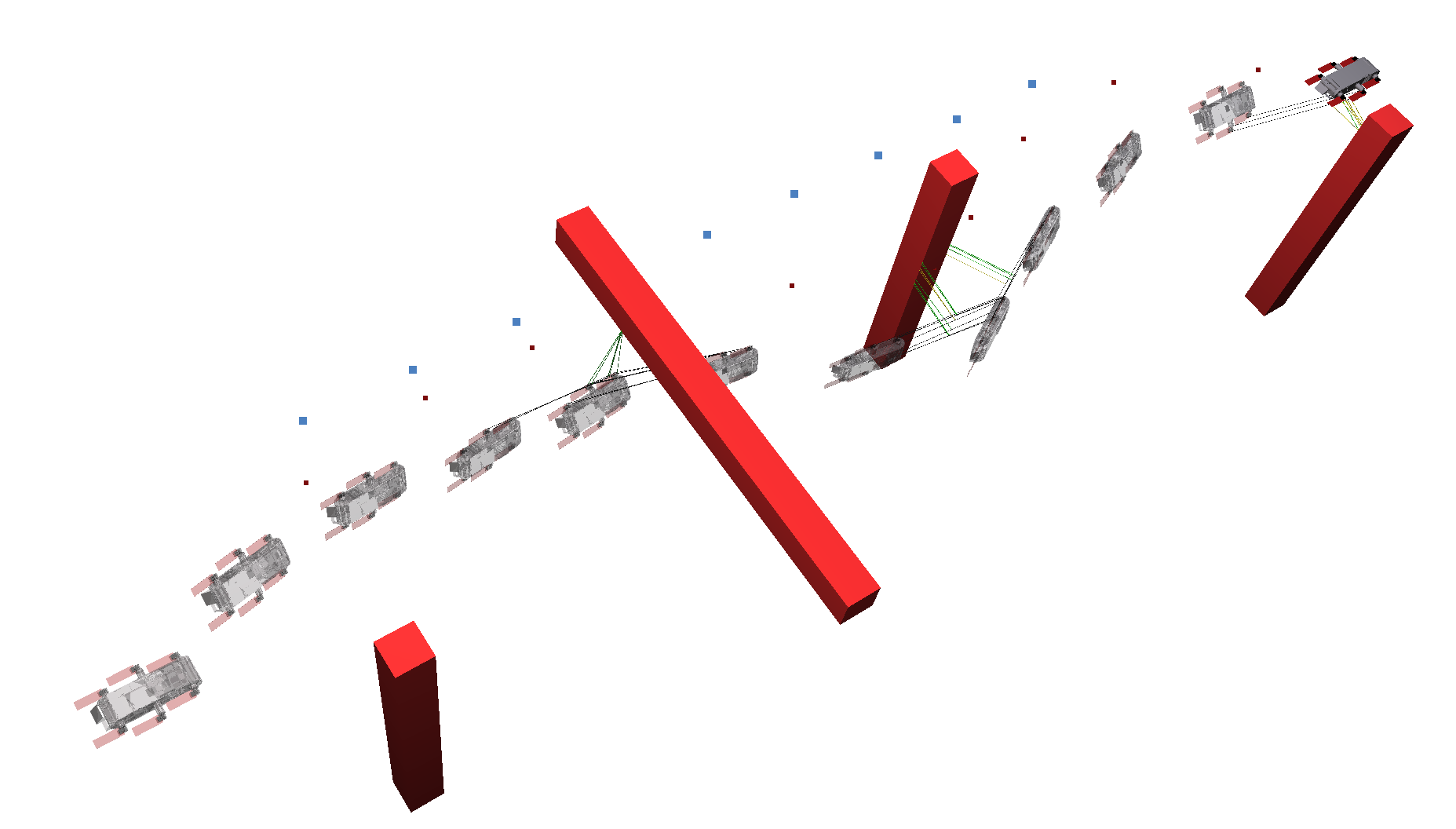}}}&
     \fbox{\subfigure[]{\includegraphics[width=0.4\columnwidth]{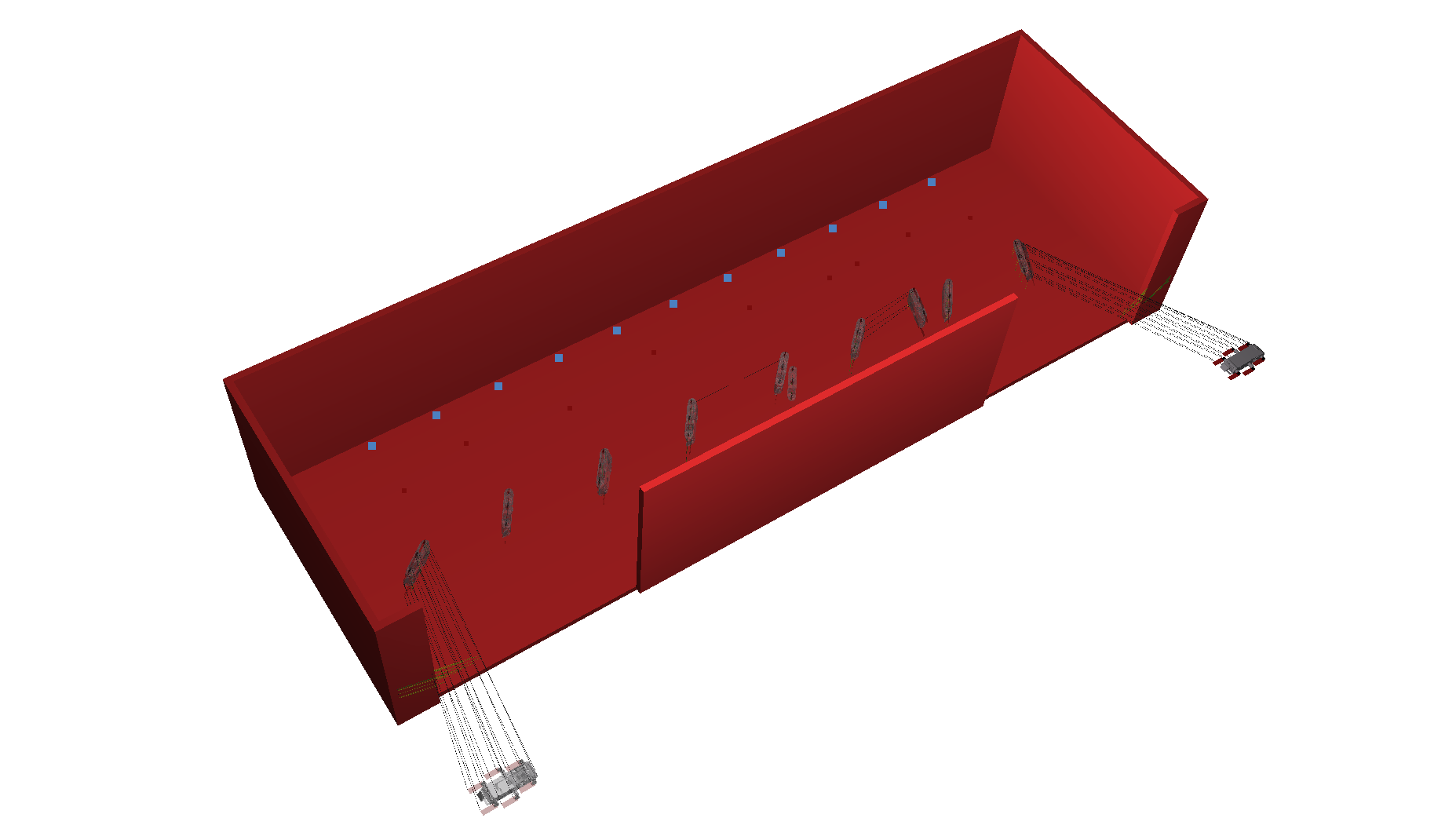}}}
    \end{tabular}
  \end{center}
\caption{Sensitivity Analysis environments. Here, obstacles are colored red and the visual objectives are denoted with blue squares}
\label{fig:Simulation}
\end{figure}

\subsection{Sensitivity Analysis}
To evaluate the flexibility of AquaVis to control the trade\hyp off between the path length, kinematics, and tracking the visual objectives, the system was tested on five simulated environments of gradually increasing complexity and difficulty. The environments considered are shown in Figure~\ref{fig:Simulation} plus one more environment with objectives as Figure~\ref{fig:Simulation}(a) but without any obstacles. The initial setup for all environments has Aqua in an straight trajectory 12m long.

There are six different weights in the cost function of AquaVis. Two weights for the initial TrajOpt formulation controlling the translation and rotation change between states, an obstacle avoidance weight, and then the two weights introduced in AquaNav to ensure the traversability of the trajectory and finally the visual objectives weight introduced in this work. Ten different weight sets were used to find their effect on the total path length and the tracking of the visual objectives.

For all environments and all sets of weights, the total trajectory length, the average distance from the nearest visual objective ($d_{obj}$) and average alignment distance ($d_{align}$) for each state $s$ were measured. A linear regression of the form $K = Q^\top W + b$ was used to quantify the effect of the weights on the parameters in question. Here $Q$ is the vector of the coefficients and $W$ is the vector that contains the base-10 logarithm of the weights. The results of the sensitivity analysis are shown in \tab{tab:analysis_results} for the coefficients of translation $t_w$, rotation  $r_w$, clearance $o_w$, way-points uniformity $d_w$, alignment  $a_w$, and visibility $v_w$.
%\vspace{-0.1in}
\begin{table}[h]
\centering
\caption{Linear Regression Coefficients}
\label{tab:analysis_results}
\begin{tabular}{|c|c|c|c|c|c|c|c|}
    \hline
    & $t_w$ & $r_w$ & $o_w$ & $d_w$ & $a_w$ & $v_w$ & b \\
    \hline
   \hspace{-0.09in}{\footnotesize Vis. Obj.}\hspace{-0.08in}  &  0.14 &  0.32 &  0.24 &  0.72 &  0.83 & -1.22 & -0.29 \\
   \hspace{-0.09in}{\footnotesize Traj. Len.}\hspace{-0.08in} & -0.50 & -0.06 &  0.22 & -0.57 & -0.32 &  0.88 & 13.48 \\
   \hspace{-0.09in}{\footnotesize Align Dis.}\hspace{-0.08in} & -0.24 & -0.03 & -0.30 & -0.15 & -0.79 &  0.81 &  2.42 \\
    \hline
\end{tabular}
\end{table}
%\vspace{-0.1in}

As expected, the only parameter that pushes the trajectory close to the visual objectives is the visual objective weight while the same parameter is the most significant in making the trajectory longer and worsening the alignment between the states. 
The trajectories shown in Figure~\ref{fig:Simulation}(a),(c) were obtained using equal weights for all parameters and thus have smooth trajectories and are moving close to the visual objectives. 
On the other hand, for the trajectories in Figure~\ref{fig:Simulation}(b) and (d), the weights for the visual objectives are increased to make the trajectory elongate and move towards the obstacles. In these cases, equally weighted parameters would cause AquaVis to miss all the visual objectives.

%\vspace{-0.1in}

\invis{
\subsection{Disclaimer}
\textcolor{red}{
Due to Covid 19 restrictions on both travel and access to the pool, in water experimental verification has been delayed. We anticipate that before the camera ready paper deadline we will have access to deploy the Aqua2 AUV for in water validation of the proposed approach. Note that both the AquaNav framework utilizing the proposed state estimation~\cite{RahmanIROS2019a} and the TrajOpt~\cite{schulman2014motion} trajectory planning strategy have been deployed in both pool and ocean settings.
}
}

\begin{figure}[!t]
 \begin{center}
  \leavevmode
   \begin{tabular}{c}
     \subfigure[]{\includegraphics[height=0.13\textheight]{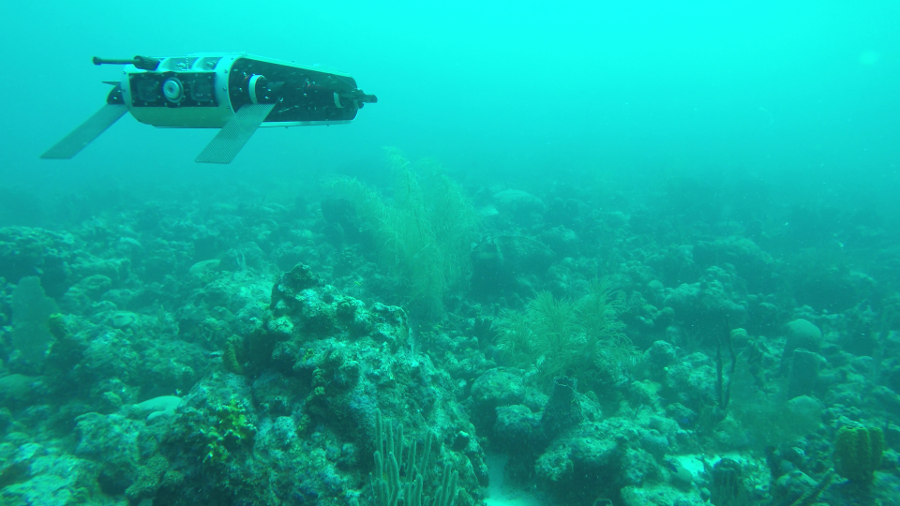}}\\
     \subfigure[]{\includegraphics[height=0.13\textheight]{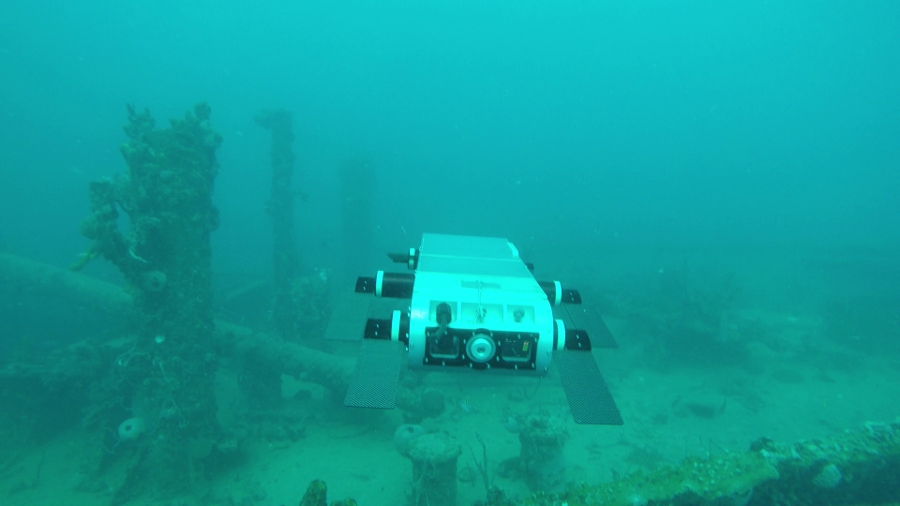}}\\
     \subfigure[]{\includegraphics[height=0.13\textheight]{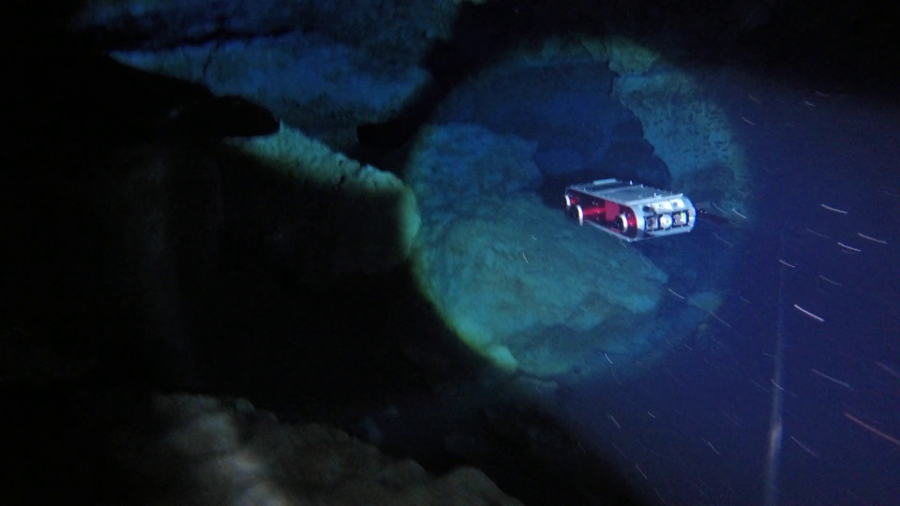}}
    \end{tabular}
  \end{center}
\caption{The Aqua2 Autonomous Underwater Vehicle operating in different environments. (a) Surveying a coral reef, mapping the corals~\cite{modasshir2019autonomous}. (b) Operating over the Stavronikita Shipwreck in Barbados. (c) Collecting data inside the Ballroom cavern in Ginnie Springs Florida.}
\label{fig:Speedo}
\end{figure}

%\vspace{-0.1in}
\section{Conclusion}
\label{concl}
%\vspace{-0.1in}
\pagebudget{0.5}
The proposed framework will enable operations of the Aqua2 vehicle in a diverse set of environments. Environmental monitoring of coral reefs will be enhanced by guiding the robot towards corals with rich features instead of sand patches; see Figure~\ref{fig:Speedo}a. Mapping underwater structures, such as shipwrecks, will benefit by ensuring the AUV operates in close proximity to the wreck and does not stray into open water where there are no features by which to localize; see Figure~\ref{fig:Speedo}b. Finally, underwater caves ---one of the most challenging environments for autonomous robots--- present additional challenges due to the restricted lighting conditions~\cite{WeidnerICRA2017}. AquaVis will guide the robot towards areas with enough light and texture to ensure safe operations; see Figure~\ref{fig:Speedo}c. 

% \afterpage{\clearpage}
%\section{PROBLEM STATEMENT}
%\input{sections/problStat}

%\input{sections/proposedMethods}

%\section{EXPERIMENTS}
%\input{sections/experiments}
%\section{CONCLUSIONS}
%%%%%%%%%%%%%%%%%%%%%%%%%%%%%%%%%%%%%%%%%%%%%%%%%%%%%%%%%%%%%%%%%%%%%%%%%%%%%%%%

\showtotalpagebudget

%%%%%%%%%%%%%%%%%%%%%%%%%%%%%%%%%%%%%%%%%%%%%%%%%%%%%%%%%%%%%%%%%%%%%%%%%%%%%%%%

%%%%%%%%%%%%%%%%%%%%%%%%%%%%%%%%%%%%%%%%%%%%%%%%%%%%%%%%%%%%%%%%%%%%%%%%%%%%%%%%
%\section*{APPENDIX}


%\section*{ACKNOWLEDGMENT}

\bibliographystyle{IEEEtran}
%\bibliography{./IEEEabrv,refs}
\bibliography{refs}

%%%%%%%%%%%%%%%%%%%%%%%%%%%%%%%%%%%%%%%%%%%%%%%%%%%%%%%%%%%%%%%%%%%%%%%%%%%%%%%%

\end{document}